\documentclass[final]{cvpr}

\usepackage{times}
\usepackage{epsfig}
\usepackage{graphicx}
\usepackage{amsmath}
\usepackage{amssymb}

\usepackage[pagebackref=true,breaklinks=true,colorlinks,bookmarks=false]{hyperref}

\usepackage{amsmath, amssymb, amsthm}
\usepackage[linesnumbered,ruled,vlined]{algorithm2e}

\SetCommentSty{mycommfont}
\SetEndCharOfAlgoLine{}
\usepackage{quoting}
\quotingsetup{vskip=0pt}
\usepackage{nicefrac}       % compact symbols for 1/2, etc.
\usepackage{bbm}
\usepackage{enumitem}
\usepackage{booktabs}       % professional-quality tables
\usepackage{appendix}

\usepackage{nccmath}
\usepackage{colortbl}
\usepackage{refcount}
\usepackage{comment}
\usepackage{caption} 
\usepackage{dsfont}

\newcommand{\update}{\textit{MemoryUpdate}}
\newcommand{\retrieval}{\textit{MemoryRetrieval}}
\newcommand\Tstrut{\rule{0pt}{2.6ex}}         % = `top' strut
\newcommand\Bstrut{\rule[-0.9ex]{0pt}{0pt}}   % = `bottom'
\usepackage{subfigure}
\usepackage{xcolor}

\def\cvprPaperID{****} % *** Enter the CVPR Paper ID here
\def\confYear{CVPR 2021}

\begin{document}

%%%%%%%%% TITLE
% \title{Supervised Contrastive Replay: for Online Class-Incremental Continual Learning}
% Supervised Contrastive Replay for Online Class-Incremental Continual Learning
\title{Supervised Contrastive Replay: Revisiting the Nearest Class Mean Classifier\\ in Online Class-Incremental Continual Learning}

\author{Zheda Mai\textsuperscript{\rm 1}, Ruiwen Li\textsuperscript{\rm 1}, Hyunwoo Kim\textsuperscript{\rm 2}, Scott Sanner\textsuperscript{\rm 1}\\
\textsuperscript{\rm 1}University of Toronto\\
\textsuperscript{\rm 2}LG AI Research\\
{\tt\small \{zheda.mai, ruiwen.li\}@mail.utoronto.ca, hwkim@lgresearch.ai, ssanner@mie.utoronto.ca}

% \and
}

\maketitle

%%%%%%%%% ABSTRACT
\begin{abstract}
% Online class-incremental continual learning (CL) studies the problem of learning new classes continually from an online non-stationary data stream, intending to adapt to new data while mitigating catastrophic forgetting—an abrupt performance degradation on previous data. 
% While memory replay techniques have shown promising results, the recency bias caused by the commonly used Softmax classifier remains an unsolved challenge. Although the Nearest-Class-Mean (NCM) classifier is significantly undervalued in the CL community, we demonstrate that it is a simple yet effective substitute for the Softmax classifier as it not only resolves several deficiencies of the Softmax classifier but also shows considerable and consistent performance gains when it's equipped on various memory replay methods. 
Online class-incremental continual learning (CL) studies the problem of learning new classes continually from an online non-stationary data stream, intending to adapt to new data while mitigating catastrophic forgetting. While memory replay has shown promising results, the recency bias in online learning caused by the commonly used Softmax classifier remains an unsolved challenge. Although the Nearest-Class-Mean (NCM) classifier is significantly undervalued in the CL community, we demonstrate that it is a simple yet effective substitute for the Softmax classifier. It addresses the recency bias and avoids structural changes in the fully-connected layer for new classes. Moreover, we observe considerable and consistent performance gains when replacing the Softmax classifier with the NCM classifier for several state-of-the-art replay methods.

To leverage the NCM classifier more effectively, data embeddings belonging to the same class should be clustered and well-separated from those with a different class label. To this end, we contribute Supervised Contrastive Replay (SCR), which explicitly encourages samples from the same class to cluster tightly in embedding space while pushing those of different classes further apart during replay-based training. Overall, we observe that our proposed SCR substantially reduces catastrophic forgetting and outperforms state-of-the-art CL methods by a significant margin on a variety of datasets.

% when replaying buffered samples with the new samples. 

%and achieves state-of-the-art performance by a significant margin on a variety of datasets.

% The Nearest-Class-Mean (NCM) classifier has shown to be a successful substitute for the Softmax classifier in continual learning to tackle recency bias and avoid the structural change in the fully connected layer for new classes. 

\end{abstract}

\section{Introduction}
With the ubiquity of personal smart devices and image-related applications, a massive amount of image data is generated daily. A practical online learning system is expected to learn incrementally without storing all  streaming data and retraining over it due to space and computational resource limitations. 
% Scott note: privacy dictates on-device training vs. collective cloud-training, which is not really hinted at here.
However, a well-documented drawback of deep neural networks that prevents it from learning continually is called \emph{catastrophic forgetting}~\cite{CF} --- the inability to retain previously learned knowledge after learning new tasks. To address this challenge, \emph{Continual Learning} (CL) studies the problem of learning from a non-i.i.d stream of data, intending to preserve and extend the acquired knowledge while minimizing storage, computation,  and time.

\begin{figure}
    \centering
    \includegraphics[width=0.9\columnwidth]{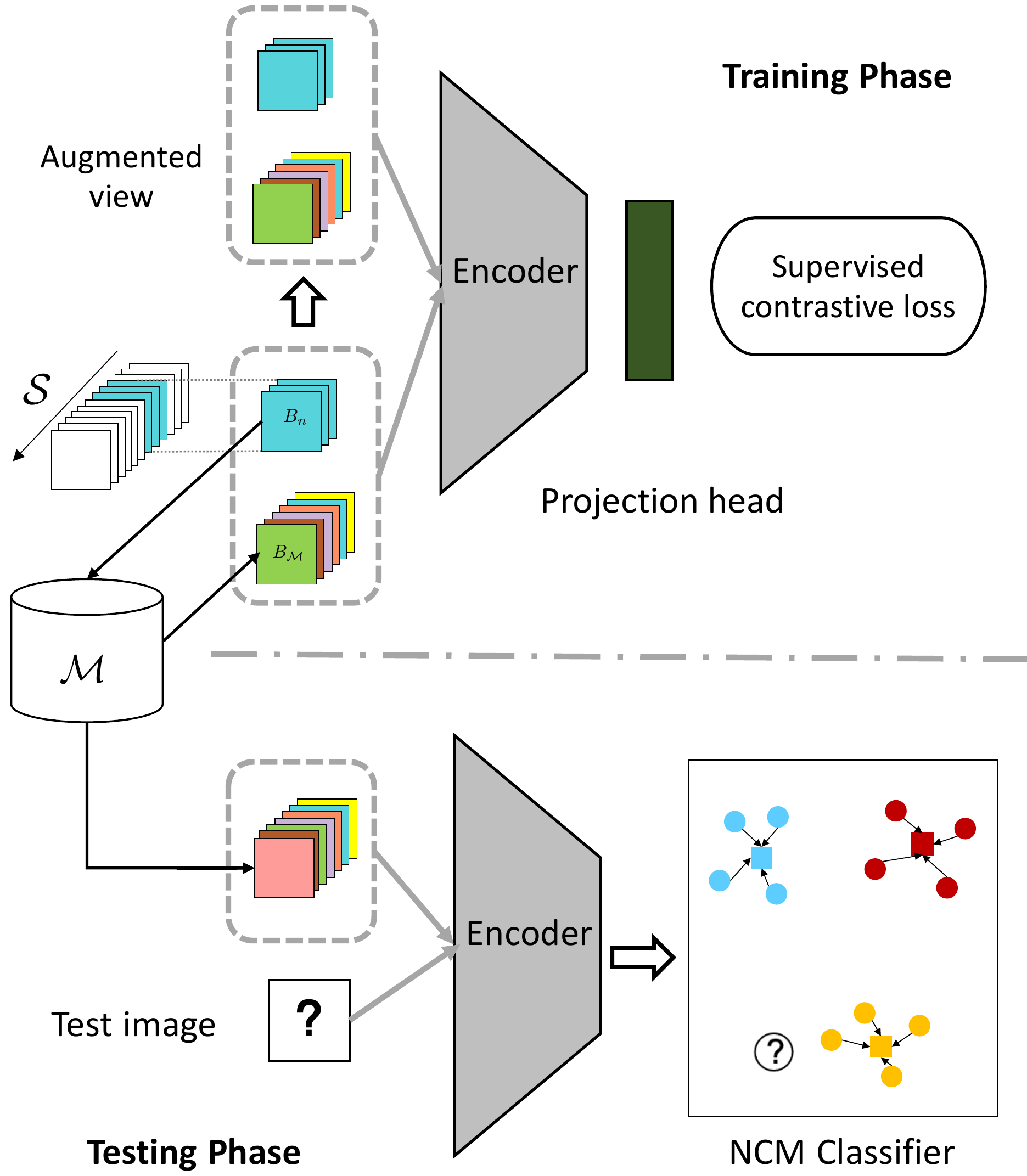}
    \caption{An overview of SCR. During training, an input batch is created by concatenating the minibatch $B_n$ from the data stream with another minibatch $B_\mathcal{M}$ from the memory buffer $\mathcal{M}$. The input batch and its augmented view are encoded by a shared encoder and projection head before the representations are evaluated by the supervised contrastive loss. During testing, the projection head is discarded and all the buffered samples are used to compute the class means for the NCM classifier.}
    \label{fig:demo}
\end{figure}
\begin{figure*}
    \centering
    \subfigure[ER]{\includegraphics[width=0.245\textwidth]{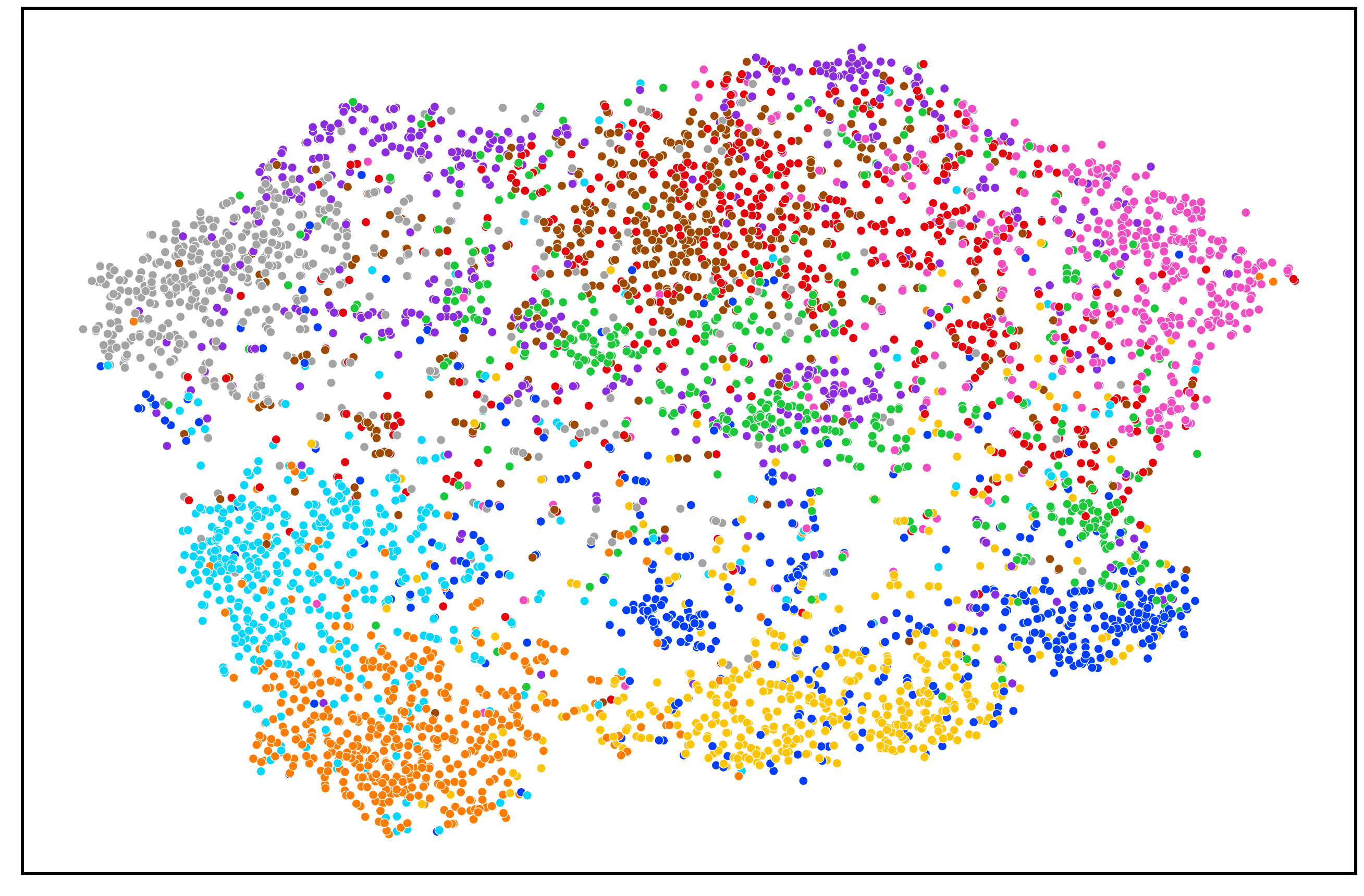}} 
    \subfigure[MIR]{\includegraphics[width=0.245\textwidth]{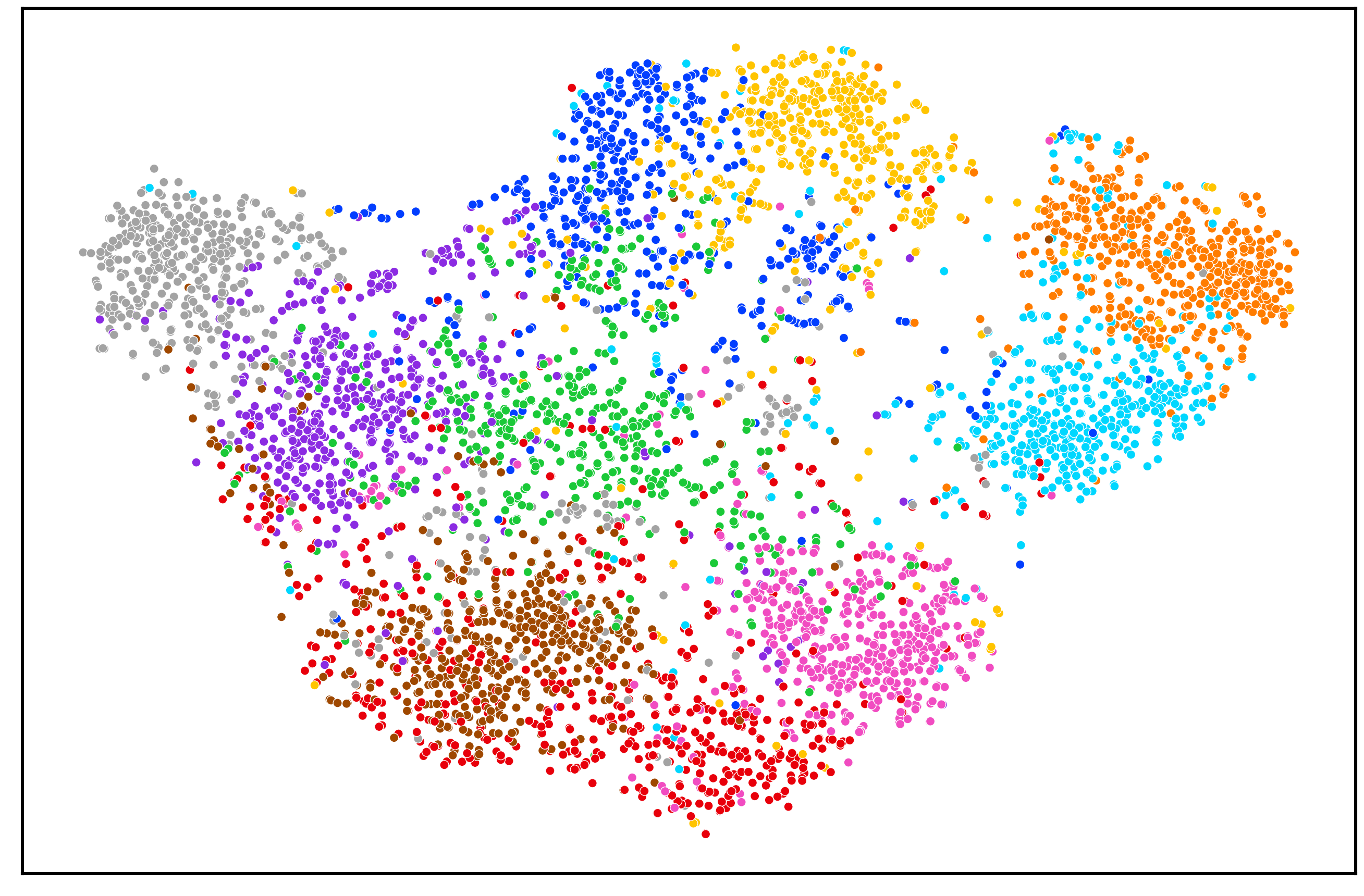}} 
    \subfigure[ASER$_\mu$]{\includegraphics[width=0.245\textwidth]{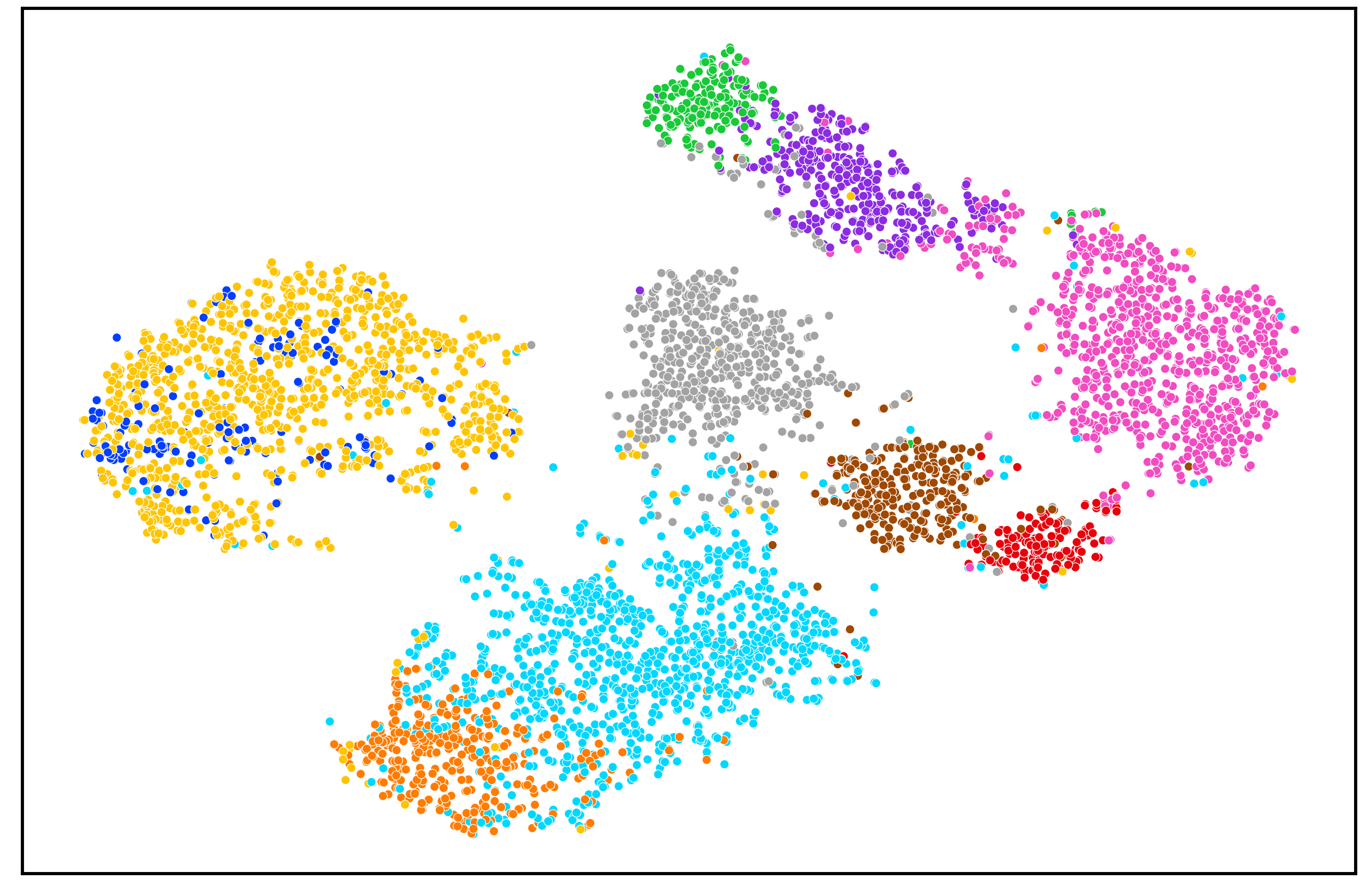}}
    \subfigure[\textbf{SCR(Ours)}]{\includegraphics[width=0.245\textwidth]{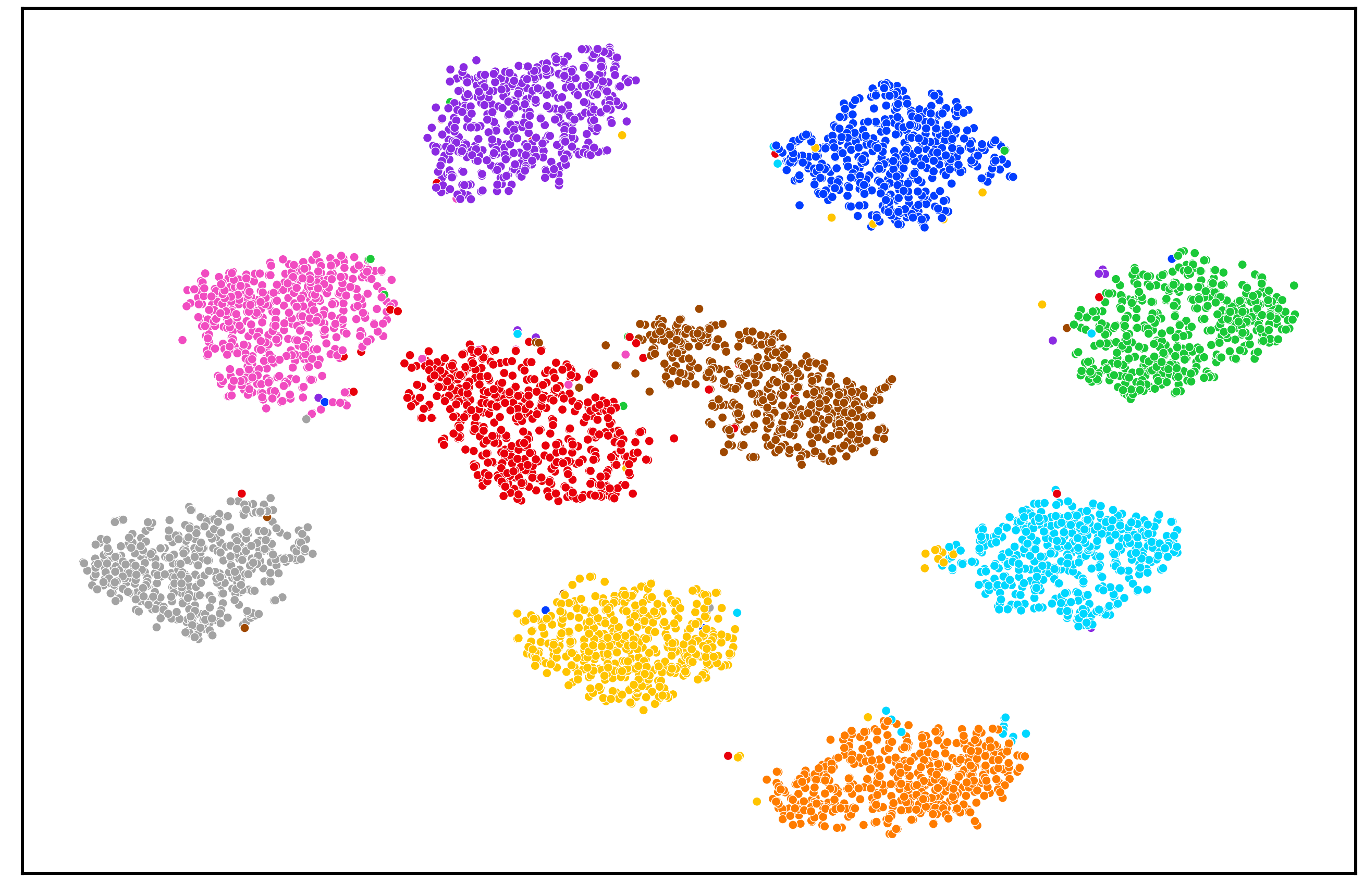}}
    % \caption{2D t-SNE \cite{tsne} visualization of data embeddings in the memory buffer ${M}$ by the end of the training (CIFAR-100). Note that ER~\cite{tiny} randomly selects samples from the buffer; MIR~\cite{mir} retrieves buffered samples that are most interfered by the incoming samples; ASER~\cite{aser} leverages Shapley value adversarially to select buffered samples. All of them use categorical cross-entropy loss to train the network. Compared with the methods mentioned above, the embeddings of our proposed SCR are better clustered and separated based on labels, which provides a solid foundation for using distance-based classifiers such as NCM~\cite{ncm} and cosine-similarity-based classifier~\cite{gidaris2018dynamic}.}
    \caption{2D t-SNE \cite{tsne} visualization of data embeddings in the memory buffer ${M}$ by the end of the training (CIFAR-100). Note that ER~\cite{tiny}, MIR~\cite{mir} and ASER~\cite{aser} are three state-of-the-art methods that use categorical cross-entropy loss to train the network. By using supervised contrastive loss, the embeddings of our proposed SCR are better clustered and separated based on labels, which provides a solid foundation for using distance-based classifiers such as NCM~\cite{ncm} and cosine-similarity-based classifier~\cite{gidaris2018dynamic}.}
    \label{fig:tsne}
\end{figure*}
Most early CL approaches considered \emph{task-incremental} settings, in which new data arrives one task at a time, and the model can utilize task-IDs during both training and inference time~\cite{Kirkpatrick2017, Li2016, gem}. This setting implicitly simplifies the CL problem as the model just needs to classify labels within a task with the help of task-IDs. Meanwhile, this simplification diminishes the applicability of this setting when task-IDs are not available. In this work, we consider a more realistic but challenging setting, known as \emph{online class-incremental}, where a model is required to learn new classes continually from an online data stream (each sample is seen only once) and classify all labels without task-IDs.

Current CL methods can be taxonomized into three major categories: regularization, parameter-isolation, and replay methods~\cite{Parisi2019, survey}. The replay approach has been shown to be simple and efficient compared to other approaches in the online class-incremental setting~\cite{mir, gss} . However, A key challenge of replay methods is the imbalance between old and new classes, as only a small amount of old class data are stored in the replay buffer. Recent works have revealed that the Softmax classifier and its associated fully-connected (FC) layer are seriously affected by the class imbalance, which leads to \emph{task-recency bias} --- the tendency of a model to be biased towards classes from the most recent task~\cite{masana2020class, mai2021online, hou2019learning, wu2019large}. Although the Nearest-Class-Mean (NCM) classifier~\cite{ncm} is significantly undervalued in the CL community, we demonstrate that it is a simple yet effective substitute for the Softmax classifier as it not only addresses the recency bias but also avoids structural changes in the FC layer when new classes are observed. Moreover, we observe considerable and consistent performance gains when replacing the Softmax classifier with the NCM classifier for five methods with memory buffers. Since \cite{yu2020semantic} also observed similar gains in methods without memory buffer, we advocate using the NCM classifier instead of the commonly used Softmax classifier for future study.

% Although several approaches have been introduced to tackle the bias~\cite{hou2019learning, wu2019large, castro2018end}, a simple yet effective approach is replacing the fully-connected layer and the Softmax classifier with the Nearest Class Mean (NCM) classifier, ~\cite{yu2020semantic, rebuffi2017icarl}.

Furthermore, to exploit the NCM classifier more effectively, the data embeddings belonging to the same class should be clustered and well-separated from those with different class labels. To this end, we contribute Supervised Contrastive Replay (SCR), which leverages the \emph{supervised contrastive loss}~\cite{khosla2020supervised} to explicitly encourage samples from the same class to cluster tightly in embedding space and push those of different classes further apart when replaying buffered samples with the new samples. Through extensive experiments on three commonly used benchmarks in the CL literature, we demonstrate that SCR outperforms state-of-the-art methods by significant margins with three different memory buffer sizes.

\section{Related Work}

\subsection{Continual Learning}
\label{sec:cl}
\paragraph{Online Class-Incremental Learning}
Following the recent CL literature \cite{mir,gss, dirichlet, onlineImbalance}, we consider the online supervised class-incremental learning setting where a model needs to learn new classes continually from an online data stream (each sample is seen only once). Formally, we define a data stream $\mathcal{D} = \{D_1, \ldots, D_N\}$ over $X\times Y$, where $X$ and $Y$ are input/output random variables respectively and $N$ is the number of tasks. Note that tasks do not overlap in classes, meaning $\{Y_i\} \cap \{Y_j\}=\emptyset$ if $i\neq j$ (where $\{Y_k\}$ represents the set of data for task $k$). We consider a classification model with two components: a encoder $f:X \mapsto\mathbb{R}^d$ that maps an input image to a compact $d$-dimensional vectorial embedding, and a classifier $g:\mathbb{R}^d \mapsto\mathbb{R}^c$ which maps the embedding to output predictions ($c$ is the number of classes observed so far). A CL algorithm $A$ is defined with the following signature:
\begin{ceqn}
\begin{align}
%$ \langle \rangle$
A_{t}:\ \langle (f, g)_{t-1}, B_t^n, M_{t-1}\rangle \ \rightarrow \ \langle (f, g)_{t}, M_{t}\rangle
\end{align}
\end{ceqn}

The model receives a small batch $B_t^n$ of size $b$ from task $D_n$ at time $t$. $f$ and $g$ will be updated based on $B_t^n$ and data in $M_{t-1}$, a bounded memory that can be used to store a subset of the training samples or other useful data~\cite{Li2016, tiny}. Moreover, we adopt the single-head evaluation setup \cite{riemannian} where the classifier has no access to task-IDs during inference and hence must choose among all labels. Our goal is to train the model $(f, g)$ to continually learn new classes from the data stream without forgetting.

\paragraph{Approaches}
As previously discussed, current CL methods can be classified into three major categories: regularization, parameter-isolation, and replay methods~\cite{Parisi2019, survey}. \textbf{Regularization} methods constraint the updates of some important network parameters to mitigate catastrophic forgetting. This is done by either incorporating additional penalty terms into the loss function~\cite{imm, mas, si, laplace} or modifying the gradient of parameters during optimization~\cite{gem, agem, he2018overcoming}. Other regularization methods imposed knowledge distillation~\cite{distill} techniques to penalize the feature drift on previous tasks~\cite{Li2016, wu2019large, encoder}. \textbf{Parameter-isolation} methods bypass interference by allocating different parameters to each task~\cite{mallya2018packnet, dirichlet, expandable}.  \textbf{Replay} methods deploy a memory buffer to store a subset of data from previous tasks for replay~\cite{mer, tiny}. Regularization methods mostly protect the model's ability to classify within a task, and thus they do not work well in our setting, which requires the ability to classify from all labels the model has seen before~\cite{lesort2019regularization}. Also, most parameter isolation methods require task-IDs during inference, which violates our setting. Therefore, in this work, we will focus on replay methods, which have been shown to be efficient and effective compared to other approaches in the online class-incremental setting~\cite{mir, gss}.

\paragraph{Metrics}
We use the \emph{average accuracy} of the test sets from observed tasks to measure the overall performance~\cite{riemannian,tiny}. In Average Accuracy, $a_{i,j}$ is the accuracy evaluated on the held-out test set of task $j$ after training the network from task 1 to $i$.  By the end of training all $N$ tasks, the average accuracy can be calculated as follows: 

\begin{align}
\text{Average Accuracy} (A_{N})&=\frac{1}{N} \sum_{j=1}^{N} a_{N, j}
\end{align}

\subsection{Contrastive Learning}

The general goal of contrastive learning is intuitive: the representation of ``similar''  samples should be mapped close together in the embedding space, while that of ``dissimilar'' samples should be further away~\cite{jaiswal2021survey, le2020contrastive}. When labels are not available (self-supervised), similar samples are often formed by data augmentations of the target sample while dissimilar samples are often drawn randomly from the same batch of the target sample~\cite{chen2020simple} or from the memory bank/queue that stores feature vectors~\cite{moco, wu2018unsupervised}. When labels are provided (supervised), similar samples are those from the same class and dissimilar samples are those from different classes~\cite{khosla2020supervised}. Contrastive learning has recently attracted a surge of interest and shown promising results in various areas including computer vision~\cite{grill2020bootstrap, caron2020unsupervised}, natural language processing~\cite{chi2020infoxlm, fang2020cert}, audio processing~\cite{schneider2019wav2vec, oord2018representation}, graph~\cite{hassani2020contrastive, qiu2020gcc} and multimodal data~\cite{arandjelovic2018objects, sun2019learning}.

\section{Method}

% \begin{figure}
%     \centering
%     \includegraphics[width=4cm]{figs/confusion.pdf}
%     \caption{[Confusion matrix of predictions from ER] [The mean of the weights in the FC layer for
% new and old classes]}
%     \label{fig:bias}
% \end{figure}

\begin{figure}
    \centering
    \subfigure[ER]{\includegraphics[width=0.49\columnwidth]{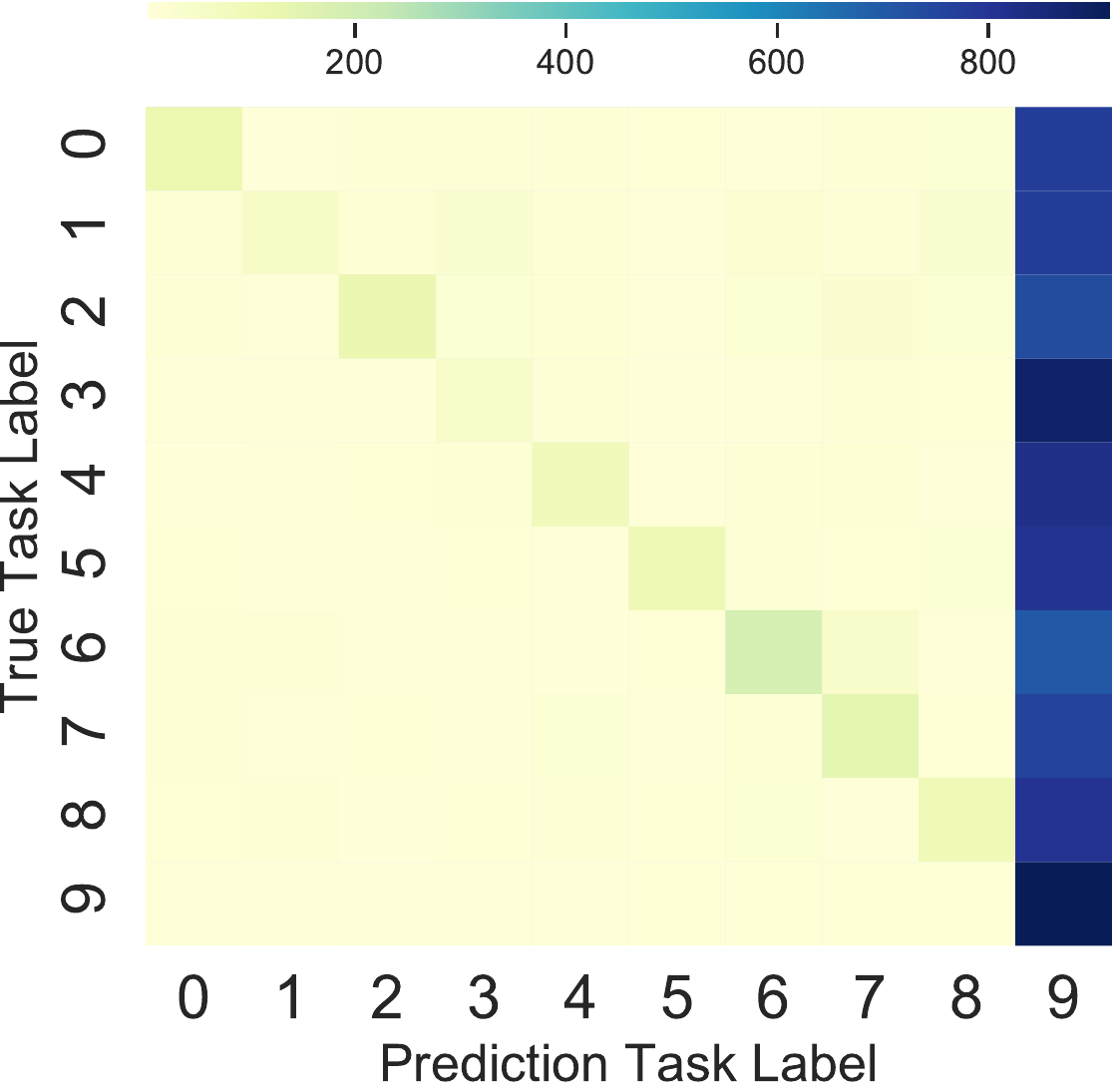}} 
    \subfigure[SCR]{\includegraphics[width=0.49\columnwidth]{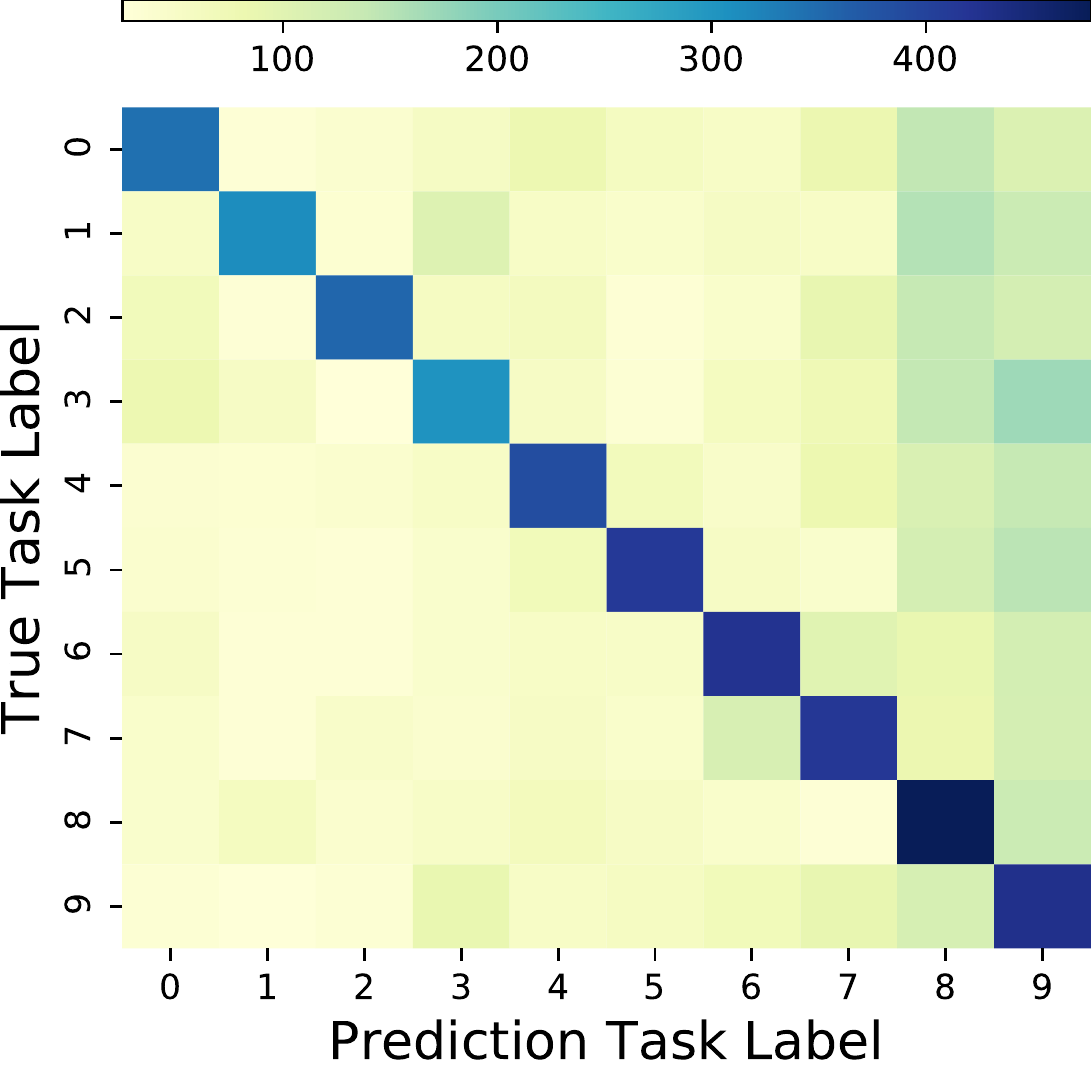}} 

    \caption{Confusion matrices for ER and SCR on CIFAR100 with a memory buffer of size 2,000. ER suffers seriously from the task-recency bias as it tends to predict most samples as classes in the most recent task, while SCR is clearly less biased because of the more discernible embeddings and NCM classifier.}
    \label{fig:confusion}
\end{figure}

\subsection{Softmax Classifier vs. NCM Classifier}
\begin{figure}
    \centering
    \includegraphics[width=6cm]{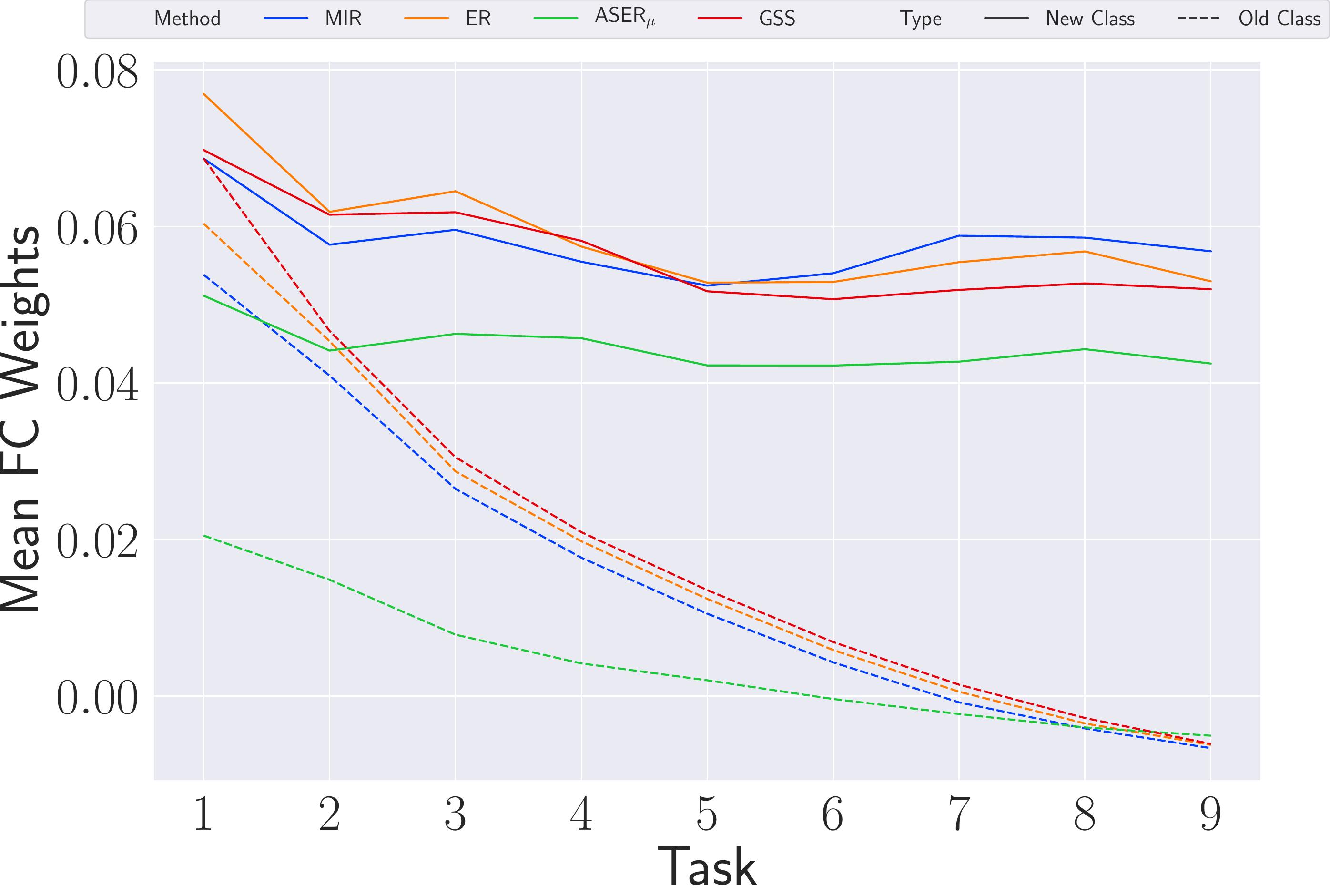}
    \caption{The means of the weights in the FC layer for
new and old classes on CIFAR100. The mean of new classes is much higher than that of old classes, which leads to task-recency bias.}
    \label{fig:fc_w}
\end{figure}

\paragraph{Softmax classifier} Softmax classifier with cross-entropy loss has been a standard approach for classification tasks for neural networks~\cite{goodfellow2016deep}. Although this combination also dominates the CL for image classification, it may not be the best choice for CL due to the following deficiencies. 
\begin{itemize}
\item \textbf{Architecture modification for new classes} When the model receives new classes, the Softmax classifier requires the model to stop training and add weights in the FC classification layer to accommodate the new classes. 
\item \textbf{Decoupled representation and classification} In the class-incremental setting, as mentioned in~\cite{rebuffi2017icarl}, it is problematic that the weights in the classification layers are decoupled from the encoder since whenever the encoder changes, weights in the classification layer must also be updated. 

\item \textbf{Task-recency bias} Multiple previous works~\cite{wu2019large, hou2019learning, lee2019overcoming, belouadah2019il2m} have observed that a model with the Softmax classifier has a strong prediction bias towards the most recent task due to the imbalance of new and old classes, which is the primary source of catastrophic forgetting. Figure~\ref{fig:confusion} (a) shows the confusion matrix after training task 10, which shows that the model tends to predict most samples as classes in the most recent task. As illustrated in Figure~\ref{fig:fc_w}, the means of weights for the new classes in the FC layer are much higher than those for the old classes and hence 
%. Since the biased weights in the FC layer have direct impacts on the output logits, 
the model assigns a larger probability mass for predicting a sample as a new class vs. an old class. 
\end{itemize}

\paragraph{Nearest Class Mean (NCM) Classifier} 

The NCM classifier and its variants have been widely used in few-shot or zero-shot learning~\cite{wang2020generalizing, xian2018zero}. Concretely, after the embedding network $f$ is trained, the NCM classifier computes a class mean (prototype) vector for each class using all the embeddings of this class. To predict a label for a new sample $\mathbf{x}$, NCM compares the embedding of $\mathbf{x}$ with all the prototypes and assigns the class label with the most similar prototype:
\begin{ceqn}
\begin{align}
\mu_{c}=\frac{1}{n_{c}} \sum_{i} f(\mathbf{x}_i)\cdot \mathbbm{1}\{\mathbf{y}_{i} = c\} \label{eq:prototye}\\
y^{*}=\underset{c=1, \ldots,
t}{\operatorname{argmin}}\left\|f(\mathbf{x})-\mu_{c}\right\|
\end{align}
\end{ceqn}
where $n_c$ is the number of samples for class $c$ and $\mathbbm{1}\{\mathbf{y}_{i} = c\}$ is the indicator for $\mathbf{y}_{i} = c$. The embedding network $f$ keeps being updated in CL, and the true prototype vector for each class cannot be exactly computed with the updated $f$ due to the unavailability of the training data for previous tasks. iCaRL~\cite{rebuffi2017icarl} approximates the prototype vectors using the data in the memory buffer, while SDC~\cite{yu2020semantic} proposes a drift compensation to update previously computed prototypes without using a memory buffer and . 

Although the NCM classifier is significantly undervalued in the CL community, we argue that it is a simple yet effective substitute for the Softmax classifier as it not only resolves the deficiencies of the Softmax classifier mentioned above but also demonstrates a considerable improvement. 

\begin{itemize}
\item Since the NCM classifier simply compares the embedding of the test sample with prototypes, it does not require an additional FC layer, and therefore, new classes can be added without any architecture modification. 

\item As the prototypes change instinctively based on the encoder, the NCM classifier is more robust against changes of the encoder. 

\item The biased weights in the FC layer result in the task-recency bias, but since the NCM classifier does not involve the FC layer, it is intrinsically less prone to the task-recency bias.
\end{itemize}

\begin{figure*}
    \centering
    \includegraphics[
    width=0.9\textwidth]{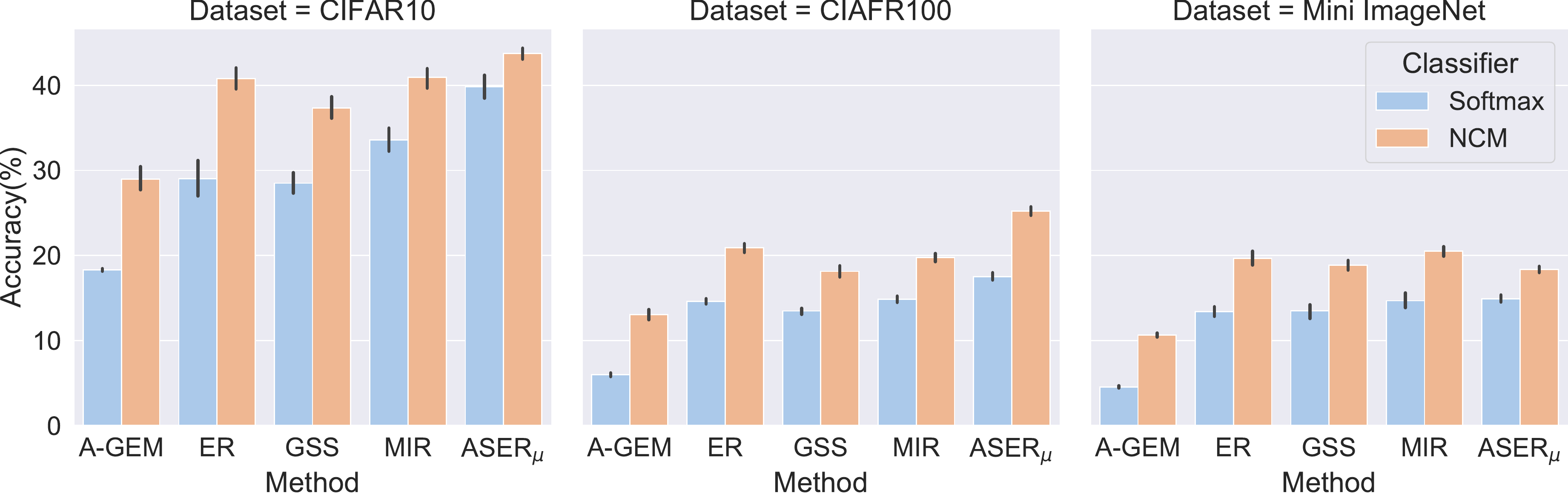}
    \caption{ The average accuracy comparison of Softmax classifier and NCM classifier on five methods with memory buffers. We set the memory size to 2,000 for Mini-ImageNet and CIFAR-100 and 500 for CIFAR-10. (Refer to Table.~\ref{tab:acc_main} for a more detailed comparison with different memory sizes). Methods with NCM classifiers show significant improvements over the commonly used Softmax classifier across all three datasets, which suggests that the dominance of the Softmax classifier in online continual learning should be revisited.}
    \label{fig:trick_cifar}
\end{figure*}

Figure ~\ref{fig:trick_cifar} shows the average accuracy comparison of a Softmax classifier and an NCM classifier on five methods. NCM classifiers show significant improvements over the commonly used Softmax classifier across all five methods and three datasets, which suggests that the dominance of the Softmax classifier in online continual learning should be revisited.
% Nice!

Although the NCM classifier has shown impressive results, the embedding quality greatly and directly impacts the performance of the NCM classifier. To effectively exploit the NCM classifier, the data embeddings belonging to the same class should be clustered and well-separated from those with a different class label. However, the binary cross-entropy loss used in iCaRL may not be capable of addressing the relationship between classes, and the commonly used categorical cross-entropy loss may not be effective in creating discernible patterns in the embedding space, as shown in Figure~\ref{fig:tsne}.

\subsection{Supervised Contrastive Replay}

\paragraph{Supervised Contrastive Learning} To improve accuracy of the vanilla NCM classifier, we propose to leverage contrastive learning, which has shown promising progress in self-supervised learning to obtain more discernible patterns in the embedding space. Specifically, we will focus on the \emph{supervised contrastive learning} (SCL)~\cite{khosla2020supervised, gunel2020supervised} as labels are available in the online class-incremental setting. Intuitively, SCL aims to tightly cluster embeddings of samples from the same class while pushing those of different classes further apart. Concretely, following the framework proposed in~\cite{chen2020simple, tian2019contrastive}, SCL consists of three main components. $Aug(\cdot)$ create an augmented view $\tilde{x}$ of a data sample $x$, $\widetilde{x}=Aug(x)$. Encoder network $Enc(\cdot)$ maps an image sample $\boldsymbol{x}$ to a vectorial embedding ${r}={Enc}({x}) \in \mathcal{R}^{D_{E}}$ (with ${r}$ normalized to the unit hypersphere in $\mathcal{R}^{D_{E}}$).Projection network $Proj(\cdot)$ maps ${r}$ to a projected vector ${z}={Proj}({r}) \in \mathcal{R}^{D_{P}}$ followed by a L2 normalization step. For an incoming batch with $b$ samples $B = \{x_k, y_k\}_{k=1 \dots b}$, we create a multiviewed batch with 2b samples: the original incoming batch and its augmented view, $B_I = B \cup \widetilde{B}$ where $\widetilde{B} = {\{\tilde{x_k}=Aug(x_k), y_k\}_{k=1 \dots b}}$. The SCL loss takes the following form:

% $\{(\boldsymbol{x}, \boldsymbol{y}), (Aug(\boldsymbol{x}), \boldsymbol{y})\}$. 

\begin{equation}
\mathcal{L}_{\text {SCL}}(Z_I)=\sum_{i \in I} \frac{-1}{|P(i)|} \sum_{p \in P(i)} \log \frac{\exp \left({z}_{i} \cdot {z}_{p} / \tau\right)}{\sum_{j \in A(i)} \exp \left({z}_{i} \cdot {z}_{j} / \tau\right)}
\label{eq: scl}
\end{equation} 

where $I$ is the set of indices of $B_I$ and $A(i)=I\backslash\{i\}$, represents the set of indices of all samples in $B_I$ except for sample $i$. $P(i) \equiv\left\{p \in A(i): {\boldsymbol{y}}_{p}={\boldsymbol{y}}_{i}\right\}$ is the set of all positives (i.e., samples with the same labels as sample $i$) in $B_I$ excluding sample $i$, and $|P(i)|$ is its cardinality. $Z_I = \{z_i\}_{i\in I}=\{{Proj}({Enc}({{x_i}})\}_{i\in I}$; $\tau \in \mathcal{R}^{+}$ is an adjustable temperature parameter controlling the separation of classes; the $\cdot$ indicates the dot product.

\paragraph{Supervised Contrastive Replay (SCR)} 
An overview of SCR can be found in Figure ~\ref{fig:demo}. As mentioned in Section~\ref{sec:cl}, during the training phase, the model receives one small batch $B_n$ at a time from task $D_n$ in the data stream $\mathcal{D}$. An input batch is created by concatenating $B_n$ with another batch $B_\mathcal{M}$ selected from the memory buffer $\mathcal{M}$. The input batch and its augmented view are encoded by a shared encoder network $Enc(\cdot)$ and a projection network $Proj(\cdot)$ before the representations are evaluated by the supervised contrastive loss $\mathcal{L}_{\text {SCL}}$. After updating both $Enc(\cdot)$
and $Proj(\cdot)$ with the gradient from $\mathcal{L}_{\text {SCL}}$, the memory buffer $\mathcal{M}$ will be updated with $B_n$. 

During the testing phase, $Proj(\cdot)$ is discarded. All the buffered samples are fed into $Enc(\cdot)$ to obtain the embeddings, which are used to compute the class means (prototypes) for the NCM classifier. As SCR builds much more discernible patterns in the embedding space with the contrastive loss, the NCM classifier is able to unleash its capability in our method. Algorithm~\ref{alg:scr} summarizes the training and inference procedures.

\begin{figure*}
    \centering
    \subfigure[CIFAR10]{\includegraphics[width=0.33\textwidth]{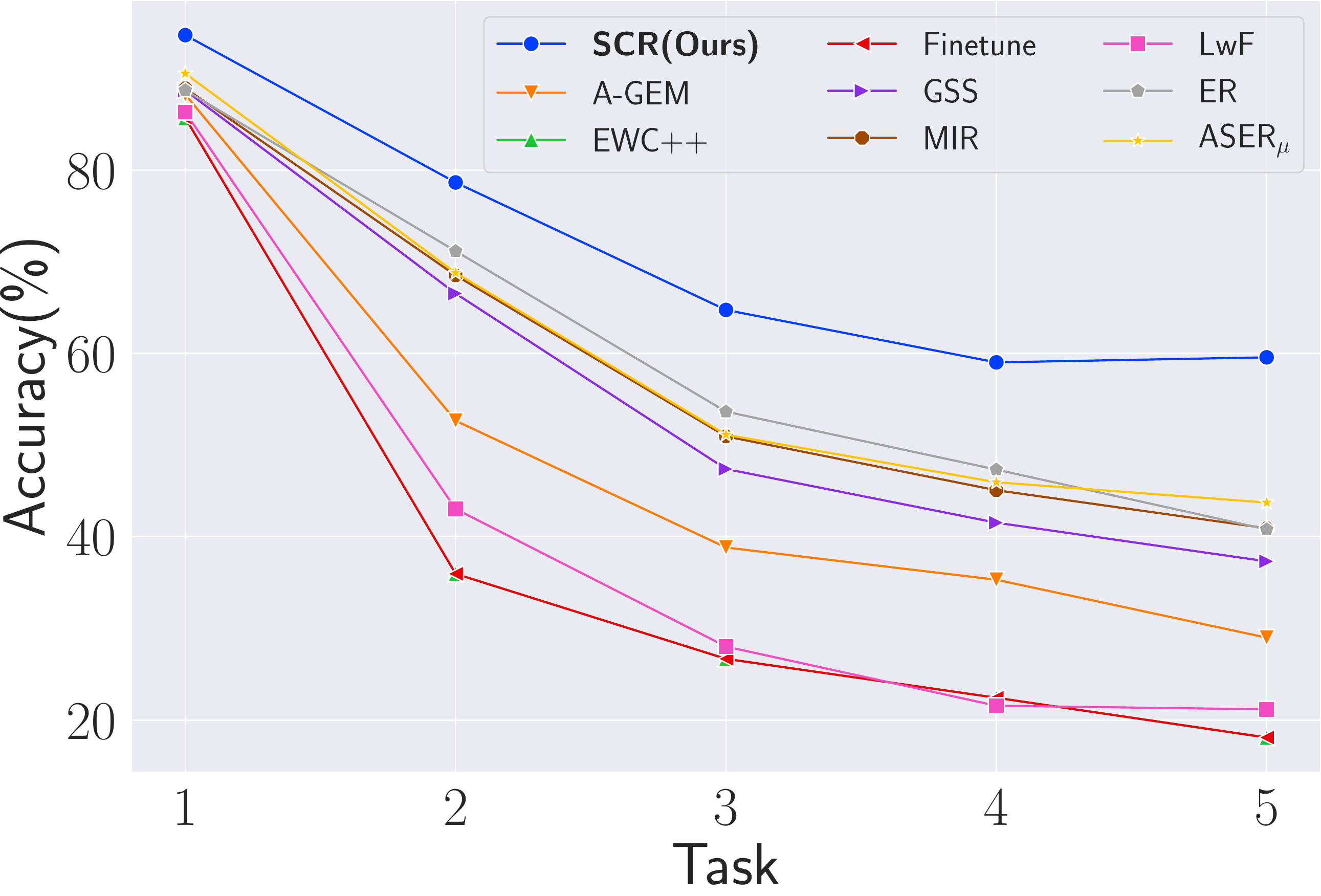}} 
    \subfigure[CIFAR100]{\includegraphics[width=0.32\textwidth]{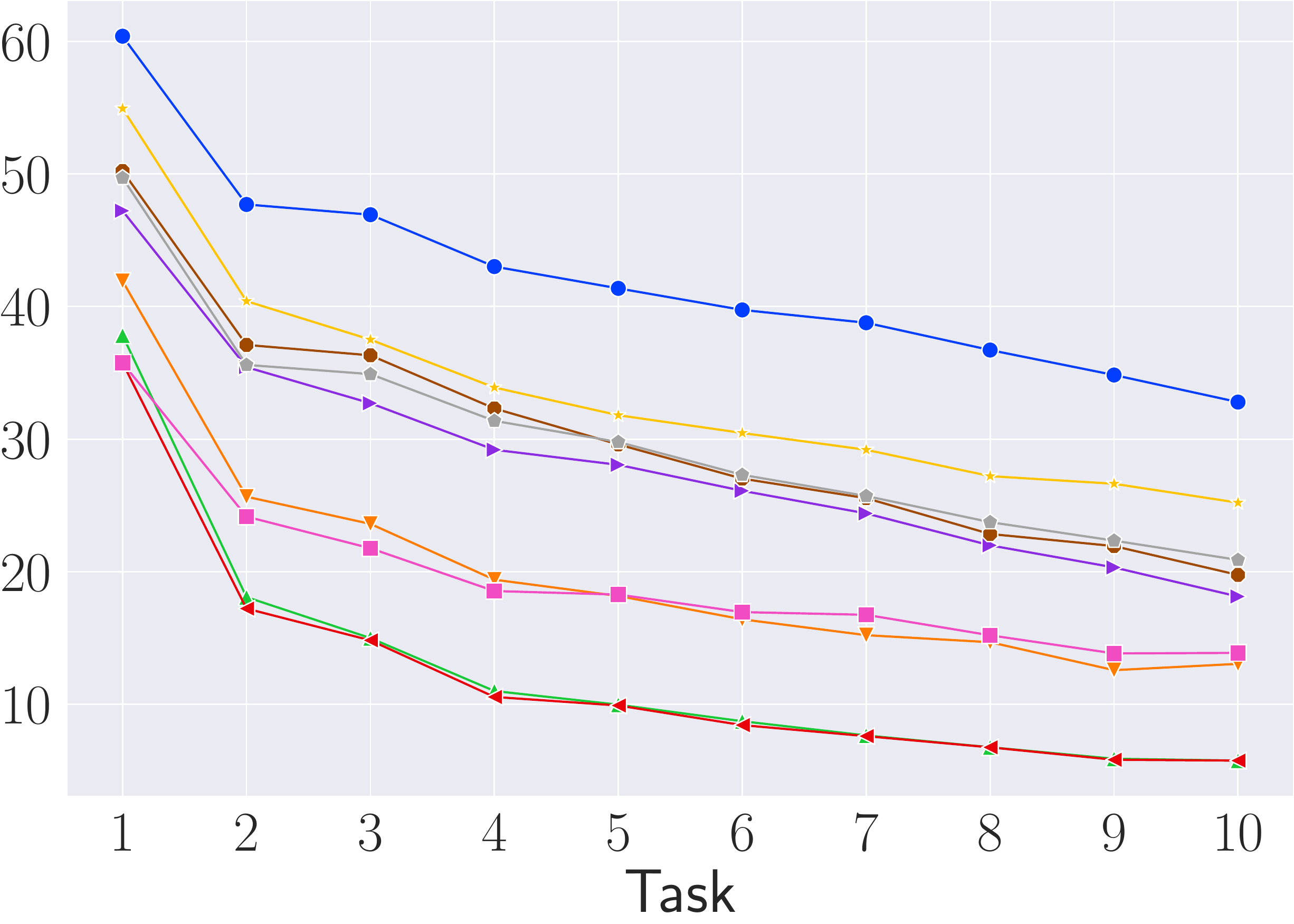}} 
    \subfigure[Mini ImageNet]{\includegraphics[width=0.32\textwidth]{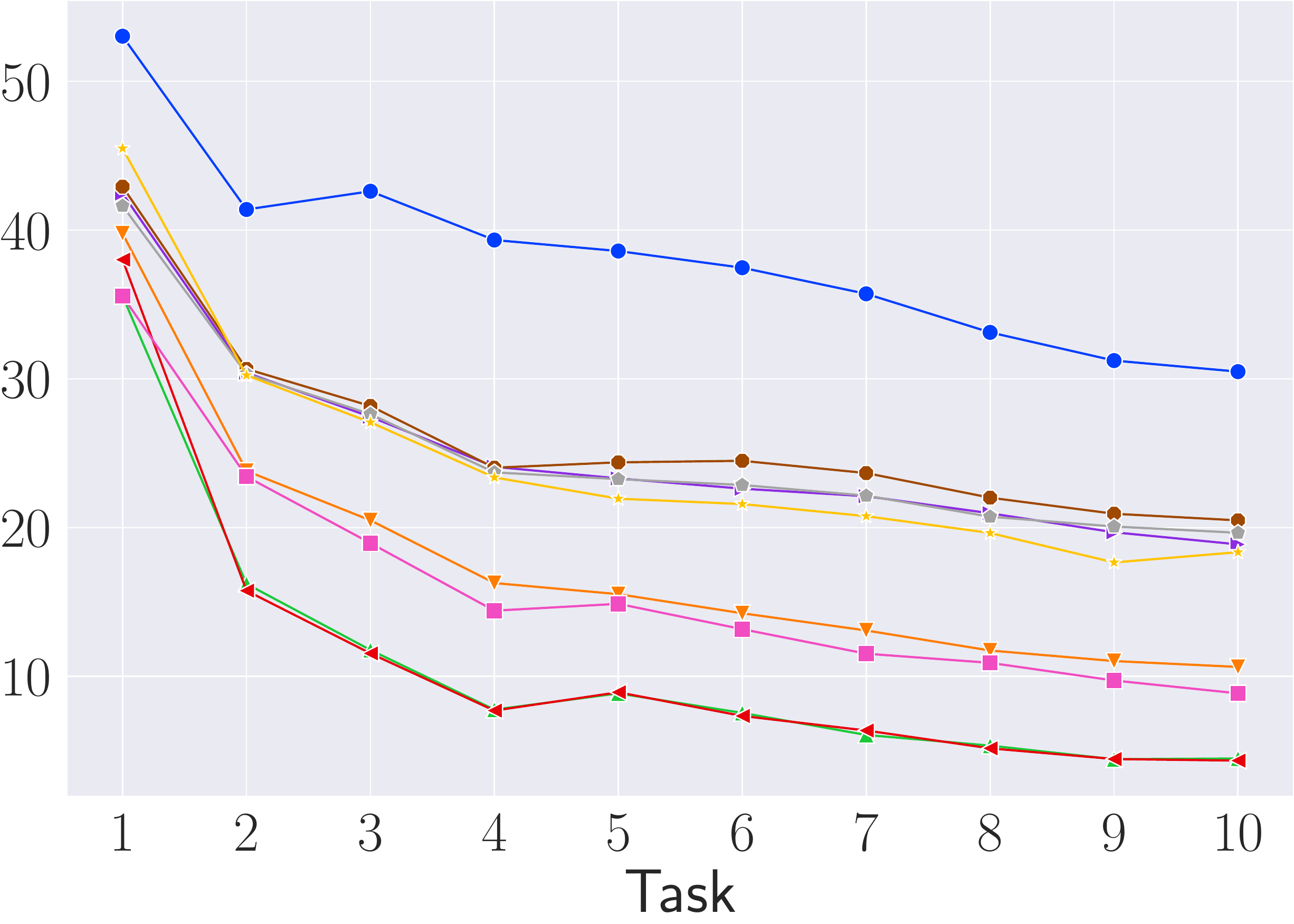}}
    \caption{Average accuracy on observed tasks on CIFAR10 (M=0.2k), CIFAR100 (M=2k) and Mini-ImageNet (M=2k). SCR consistently outperform all the compared methods by an enormous margin. Note that all the compared methods on the plots use the NCM classifier.  }
    \label{fig:line}
\end{figure*}

\begin{algorithm}
\LinesNumberedHidden
\caption{Supervised Contrastive Replay}
\label{alg:scr}
\SetCustomAlgoRuledWidth{0.49\textwidth}
\SetAlgorithmName{Algorithm}{}{}
    \SetKwInOut{Input}{Input~}
    \SetKwInOut{Output}{Output}
    \SetKwInput{Initialize}{Initialize}
% \Input{Batch size $b$, Learning rate $\alpha$}
\Initialize{Memory $\mathcal{M}\leftarrow \{\}*M$; $Aug(\cdot)$; $Enc_\theta(\cdot)$;  $Proj_\phi(\cdot)$ }

\For{$n\in\{1,\dots,N\}$}{
    \;
    \emph{Training phase:}\\
    \For{$B_n\sim D_n$}{
    $B_\mathcal{M}\!\!\leftarrow\!$ \retrieval{}($B_n,\! \mathcal{M}$)\\
    $B_{n\mathcal{M}} \leftarrow B_n\cup B_\mathcal{M}$ \\
    $B_I \leftarrow B_{n\mathcal{M}} \cup Aug( B_{n\mathcal{M}})$\\
    % $\theta\leftarrow~\text{SGD}(B_n\cup B_\mathcal{M},\theta, \alpha)$ \\
    ${Z_I} \leftarrow Proj_\phi(Enc_\theta(B_{I}))$ \\
    % ${Z_B} \leftarrow Proj_\phi(Enc_\theta(B_{n\mathcal{M}}))$ \\
    % ${Z_V} \leftarrow Proj_\phi(Enc_\theta(Aug(B_{n\mathcal{M}})))$ \\
    $\theta, \phi \leftarrow \text{SGD}(\mathcal{L}_{\text{SCL}}({Z_I}), \theta, \phi )$\tcp{Eq.~\ref{eq: scl}}
    $\mathcal{M}\leftarrow$ \update{}$(B_n, \mathcal{M})$\\ 
    }
    \;
    \emph{Testing phase:}\\
    \tcp*[h]{$C$ $\leftarrow$ number of observed classes }\\
    \For{$c\in\{1,\dots,C\}$}{ 
    \tcp*[h]{$n_c$ $\leftarrow$ number of class $c$ samples}
    $\mu_{c}=\frac{1}{n_{c}} \sum\limits_{i}^{|\mathcal{M}|} Enc_\theta({x}_i)\cdot \mathbbm{1}\{\mathbf{y}_{i} = c\}$\\
    }
    
    $y^{*}=\underset{c=1, \ldots,
    t}{\operatorname{argmin}}\left\|Enc_\theta(\mathbf{x})-\mu_{c}\right\|$ \tcp{classify $\mathbf{x}$}
}
\end{algorithm}

\section{Experiment}

% To test the efficacy of the NCM classifier and our proposed SCR, we compare five methods that employ memory buffers with their variants equipped with the NCM classifier. Moreover, we evaluate the performance of SCR by comparing them with several state-of-the-art CL methods that are NCM-augmented. We begin by reviewing the benchmark datasets, baselines we compared against and our experiment setting. We then report and analyze the result to validate our approach followed by the ablation study.

\subsection{Experiment Setup}
\label{sec:setup}
\paragraph{Datasets}\textbf{Split CIFAR-10} is constructed by splitting the CIFAR-10 dataset \cite{cifar} into 5 different tasks with non-overlapping classes and 2 classes in each task, similarly as in \cite{mir}. \textbf{Split CIFAR-100} splits the CIFAR-100 dataset \cite{cifar} into 10 disjoint tasks, and each task has 10 classes. \textbf{Split Mini-ImageNet} divides the Mini-ImageNet dataset \cite{miniimagenet} into 10 disjoint tasks with 10 classes per task.

% Within each task, there are 2 classes with 9000 samples for training, 1000 for validation and 1000 for testing.

%We consider the setting with 4500 samples for training, 500 validation samples and per task.

%with 4800 training samples and 600 validation sample. 

% The detail of datasets, including the general information of each dataset,  class composition and the number of samples in training, validation and test sets of each task is presented in Appendix A.

\paragraph{Baselines}
We compare our proposed SCR against several state-of-the-art continual learning algorithms:
\begin{itemize}
  \item\textbf{A-GEM} (ICLR'19)~\cite{agem}: Averaged Gradient Episodic Memory, that utilizes the samples in the memory buffer to constrain the parameter updates.  
  \item \textbf{ASER$_\mu$} (AAAI'21)~\cite{aser}: Adversarial Shapley Value Experience Replay that leverages Shapley value adversarially in memory retrieval. 
  \item \textbf{ER} (ICML-W'19)\cite{tiny}: Experience replay, a replay method with random sampling in memory retrieval and reservoir sampling in memory update.
    \item \textbf{EWC++} (ECCV'18)~\cite{riemannian}: An online version of EWC~\cite{Kirkpatrick2017}, a regularization method that limits the update of parameters that were crucial to the past tasks. 
    
  \item \textbf{GSS} (NeurIPS'19)~\cite{gss}: Gradient-Based Sample Selection, a replay  method that diversifies the gradients of the samples in the replay memory.
  
  \item \textbf{LwF} (TPAMI’18)~\cite{Li2016} Learning Without Forgetting, a regularization method that utilizes knowledge distillation to penalize the feature drifts on previous  tasks. 
  
  \item \textbf{MIR} (NeurIPS'19)~\cite{mir}: Maximally Interfered Retrieval, a replay method that retrieves memory samples with loss increases given the estimated parameter update based on the current batch.
  \item \textbf{offline}: This is not a CL method, but rather an upper bound; offline trains the model over multiple epochs on the whole dataset with iid sampled mini-batches. We use 50 epochs for offline training. 
  \item \textbf{fine-tune}: A lower-bound method that simply trains the model when new data is presented without any measure for forgetting avoidance.

\end{itemize}

\paragraph{Implementation Detail}
Following \cite{tiny, gem, agem, mir}, we use a reduced ResNet18 as the backbone model for all datasets. We use stochastic gradient descent with a learning rate of 0.1, and the model receives a batch with size 10 at a time from the data stream. All the methods except for SCR are trained with cross-entropy loss and classify with the Softmax classifier. The projection network of SCR is a Multi-Layer Perceptron (MLP)~\cite{goodfellow2016deep} with one hidden layer (ReLU) and an output size 128, and we set the temperature $\tau$ to 0.1. We use reservoir sampling~\cite{vitter1985random} for memory update and random sampling for memory retrieval and use a memory batch size 100. The ablation study of the variables mentioned above will be discussed in Section.~\ref{sec: ablation}. 
{
\begin{table*}[h]
    \small
    \centering
    \begin{tabular}{ c@{\kern0.7em} @{}c@{\kern0.7em} @{}c@{\kern0.7em} @{}c@{\kern0.7em} | @{\kern0.7em}c@{\kern0.7em} @{}c@{\kern0.7em} @{}c@{\kern0.7em} | @{\kern0.7em}c@{\kern0.7em} @{}c@{\kern0.7em} @{}c@{\kern0.7em} @{}c} 
    \toprule
    Method &M=1k& M=2k &  M=5k\Bstrut & M=1k& M=2k &  M=5k &  M=0.2k &  M=0.5k & M=1k \\
    % \midrule
    \hline\hline
    \Tstrut
    fine-tune &\multicolumn{3}{c}{$4.3\pm0.2$}&\multicolumn{3}{c}{$5.8\pm0.3$}& \multicolumn{3}{c}{$18.1\pm0.3$}\\
    iid offline  & \multicolumn{3}{c}{$51.4\pm0.2$}&\multicolumn{3}{c}{$49.6\pm0.2$}&\multicolumn{3}{c}{$81.7\pm0.1$}\\
    EWC++ & \multicolumn{3}{c}{$4.5\pm0.2$}&\multicolumn{3}{c}{5.8$\pm$0.3}&\multicolumn{3}{c}{$18.1\pm0.3$}\\
    LwF & \multicolumn{3}{c}{8.9$\pm$0.5}&\multicolumn{3}{c}{13.9$\pm$0.5}&\multicolumn{3}{c}{$21.2\pm0.9$}\\
    \midrule
    AGEM &$4.5\pm0.4$&$4.6\pm0.2$&$4.6\pm0.2$&$5.8\pm0.3$&$6.0\pm0.3$&$5.9\pm0.2$&$18.2\pm0.3$&$18.3\pm0.2$&$18.2\pm0.2$\\
    AGEM-NCM &$9.5\pm0.3$&$10.6\pm0.3$&$11.6\pm0.5$&$11.5\pm0.8$&$13.1\pm0.8$&$14.3\pm0.4$&$28.1\pm1.8$&$29.0\pm1.8$&$29.1\pm0.9$\\
    \midrule
    ER &$10.3\pm0.7$&$13.4\pm0.7$&$16.4\pm1.5$&$11.2\pm0.6$&$14.6\pm0.4$&$21.0\pm0.9$&$22.4\pm1.1$&$29.0\pm2.5$&$37.7\pm2.0$\\
    ER-NCM &$16.8\pm0.8$&$19.7\pm1.0$&$21.1\pm0.8$&$16.8\pm0.5$&$20.9\pm0.6$&$28.3\pm1.0$&$30.8\pm2.0$&$40.8\pm1.5$&$49.4\pm0.9$\\
    \midrule
    GSS &$10.5\pm0.6$&$13.5\pm1.1$&$14.5\pm2.2$&$10.6\pm0.4$&$13.5\pm0.4$&$18.0\pm1.1$&$23.0\pm0.9$&$28.5\pm1.5$&$34.6\pm2.3$\\
    GSS-NCM &$15.2\pm0.9$&$18.9\pm0.7$&$20.9\pm1.3$&$13.2\pm0.7$&$18.1\pm0.9$&$25.8\pm0.7$&$28.5\pm1.2$&$37.3\pm1.6$&$46.6\pm2.0$\\
    \midrule
    MIR &$10.7\pm0.7$&$14.7\pm1.1$&$17.3\pm1.6$&$11.7\pm0.3$&$14.9\pm0.5$&$21.6\pm1.2$&$23.8\pm0.9$&$33.6\pm1.7$&$43.0\pm1.6$\\
    MIR-NCM &$17.8\pm0.5$&$20.5\pm0.7$&$22.1\pm0.9$&$16.4\pm0.4$&$19.8\pm0.6$&$27.9\pm1.0$&$31.2\pm1.5$&$40.9\pm1.5$&$49.9\pm1.0$\\
    \midrule
    ASER$_\mu$&$12.5\pm0.8$&$14.9\pm0.5$&$18.2\pm0.9$&$14.4\pm0.6$&$17.5\pm0.6$&$21.7\pm1.0$&$28.5\pm1.3$&$39.8\pm1.7$&$46.7\pm1.3$\\
    ASER$_\mu$-NCM &$16.6\pm0.7$&$18.4\pm0.5$&$21.1\pm0.3$&$22.0\pm0.6$&$25.2\pm0.7$&$29.6\pm0.4$&$34.1\pm0.8$&$43.7\pm0.8$&$50.3\pm0.9$\\
    \midrule
    SCR&$\mathbf{24.1\pm0.6}$&$\mathbf{30.6\pm0.5}$&$\mathbf{35.4\pm0.5}$&$\mathbf{26.6\pm0.5}$&$\mathbf{32.8\pm0.7}$&$\mathbf{37.8\pm0.3}$&$\mathbf{48.6\pm1.1}$&$\mathbf{59.6\pm1.2}$&$\mathbf{65.7\pm0.6}$\\
    Gains&$\textcolor{red}{6.3} \; \textcolor{red}\uparrow$&$\textcolor{red}{10.1} \; \textcolor{red}\uparrow$&$\textcolor{red}{13.3} \; \textcolor{red}\uparrow$&$\textcolor{red}{4.6} \; \textcolor{red}\uparrow$&$\textcolor{red}{7.6} \; \textcolor{red}\uparrow$&$\textcolor{red}{8.2} \; \textcolor{red}\uparrow$&$\textcolor{red}{14.5} \; \textcolor{red}\uparrow$&$\textcolor{red}{15.9} \; \textcolor{red}\uparrow$&$\textcolor{red}{15.4} \; \textcolor{red}\uparrow$\\
    \hline
    \bottomrule
    \Tstrut
     &  \multicolumn{3}{c}{(a) Mini-ImageNet\label{accu-mini-imagenet} } & \multicolumn{3}{c}{\label{accu-CIFAR-100} (b) CIFAR-100} & \multicolumn{3}{c}{\label{accu-cifar-10} (c) CIFAR-10}\\
    \end{tabular}
    % \subcaption{Mini-ImageNet}
\caption{Average Accuracy by the end of training. M is the memory buffer size and all numbers are the average of 10 runs. SCR considerably and consistently outperforms all the compared methods by large margins in different datasets and memory sizes.}
\label{tab:acc_main}
\end{table*}
}

\begin{figure}
    \centering
    \includegraphics[width=5cm]{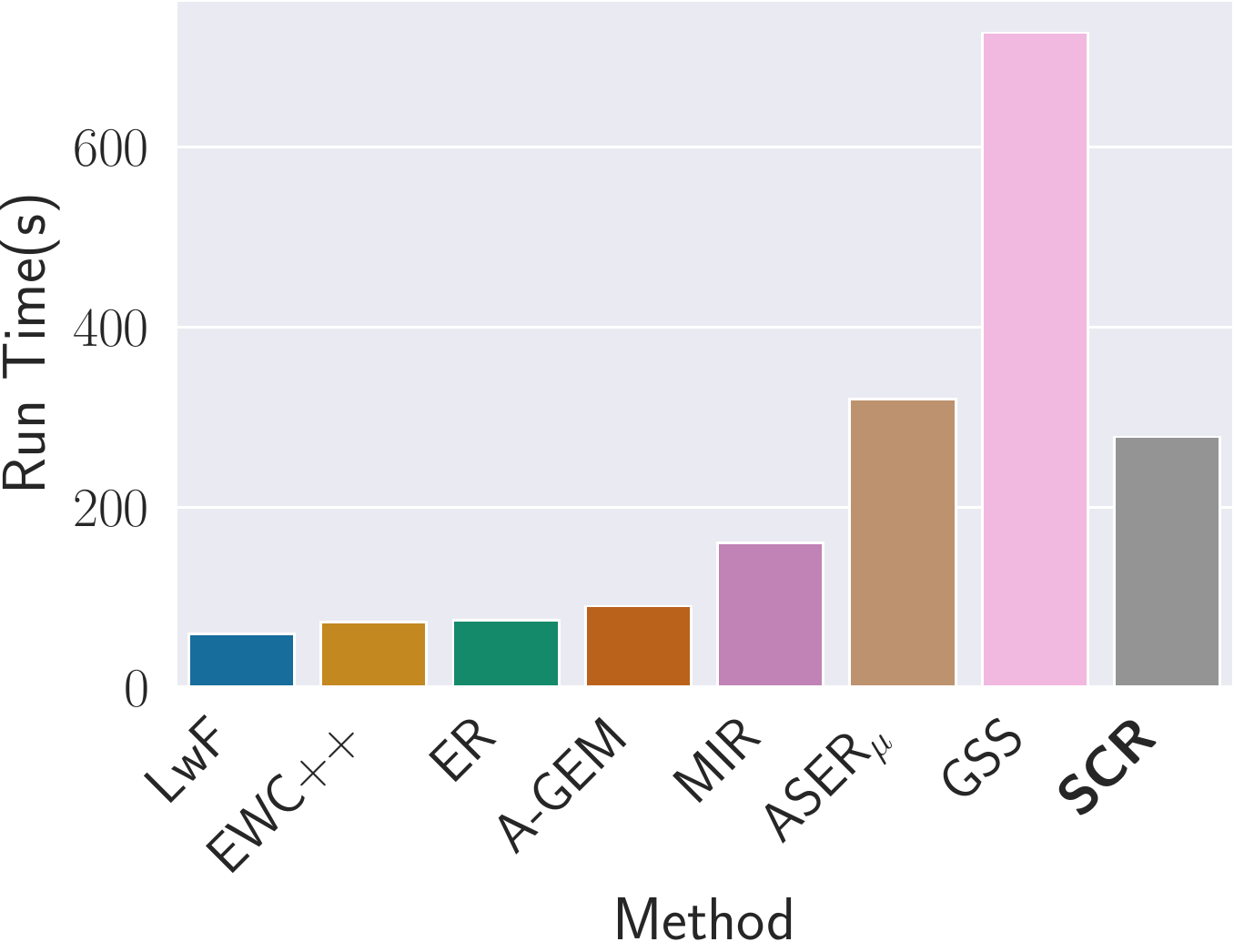}
    \caption{Run time (training + inference) comparison. SCR achieves state-of-the-art performance without sacrificing computation efficiency.}
    \label{fig:run_time}
\end{figure}

\subsection{Evaluation of NCM Classifier}
\label{sec: ncm_eval}
To assess the effectiveness of the NCM classifier, we compare five methods that employ memory buffers (AGEM, ER, GSS, MIR, ASER$_\mu$) with their variants equipped with the NCM classifier. As we can see in Figure~\ref{fig:trick_cifar} and Table~\ref{tab:acc_main}, methods with the NCM classifier show significant improvements over those with the default Softmax classifier. For instance, in CIFAR100, the NCM classifier helps ASER$_\mu$ with 1k memory achieve 22\%, which requires \emph{five times more memory} to achieve when using the Softmax classifier.  Generally, we also observe that the performance gain is more notable when the memory buffer is small. For example, in Mini-ImageNet, MIR obtains 66.4\% relative improvement (10.3\% $\rightarrow$ 17.8\%) with M=1k, which is only improved by 27.7\% relatively (17.3\% $\rightarrow$ 22.1\%) with M=5k. Furthermore, the NCM gains are less obvious for GSS, and we find out that it's because some classes only have a few or sometimes zero samples in the GSS buffer, which makes it hard to estimate the correct prototypes for those classes. Moreover, ASER$_\mu$ has better NCM gains in general, and it's because ASER$_\mu$ tends to learn more discernible embeddings, as we can see in Figure ~\ref{fig:demo}. 

To sum up, we observe considerable and consistent performance gains when replacing the Softmax classifier with the NCM classifier for five methods on three different datasets and memory sizes. Since \cite{yu2020semantic} also observed similar gains in methods without memory buffer, we advocate using the NCM classifier instead of the commonly used Softmax classifier for future study.

\begin{figure*}[ht!]
    \centering
    \subfigure[Impact of memory batch size]{\includegraphics[width=0.24\textwidth]{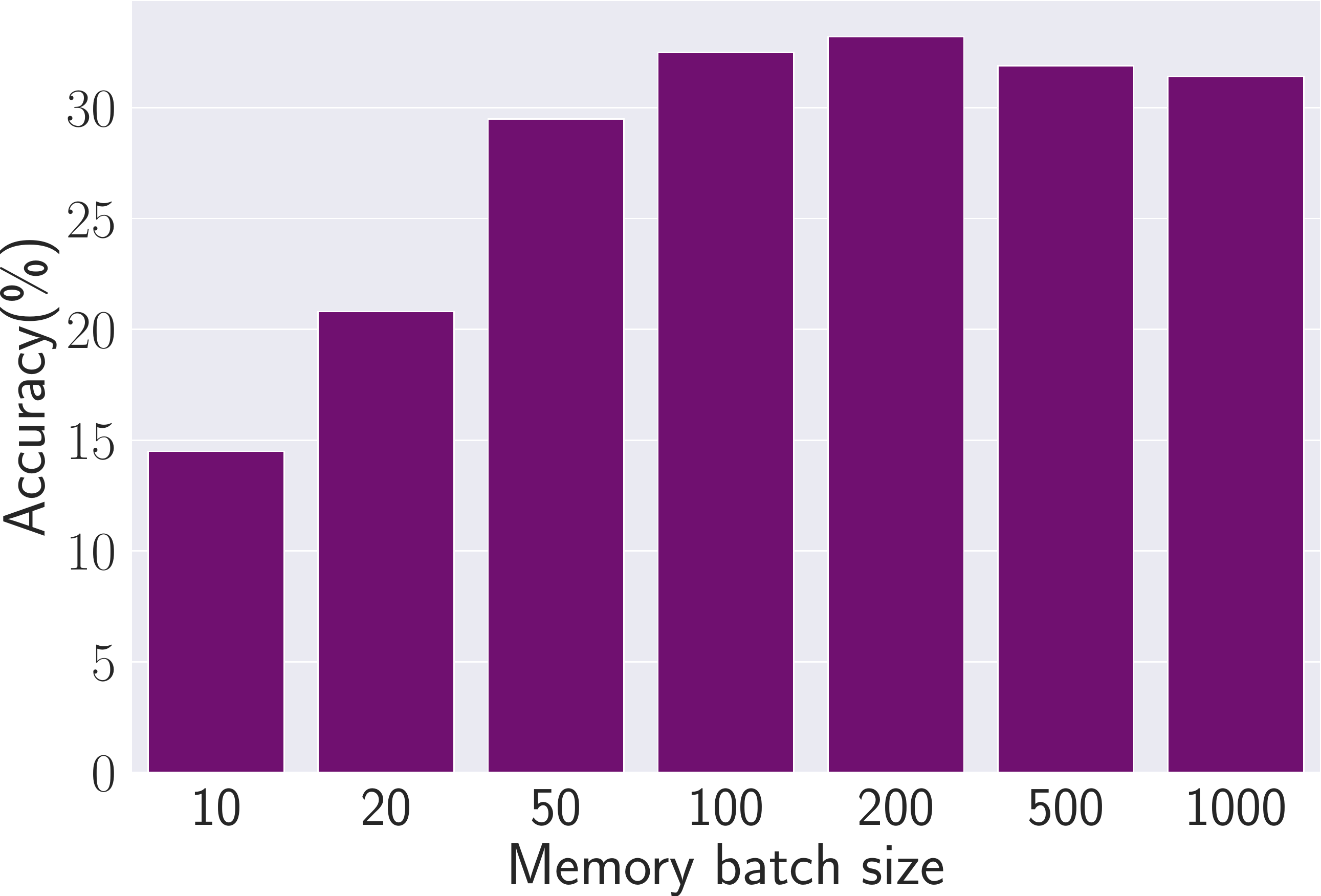}} 
    \subfigure[Impact of memory management]{\includegraphics[width=0.24\textwidth]{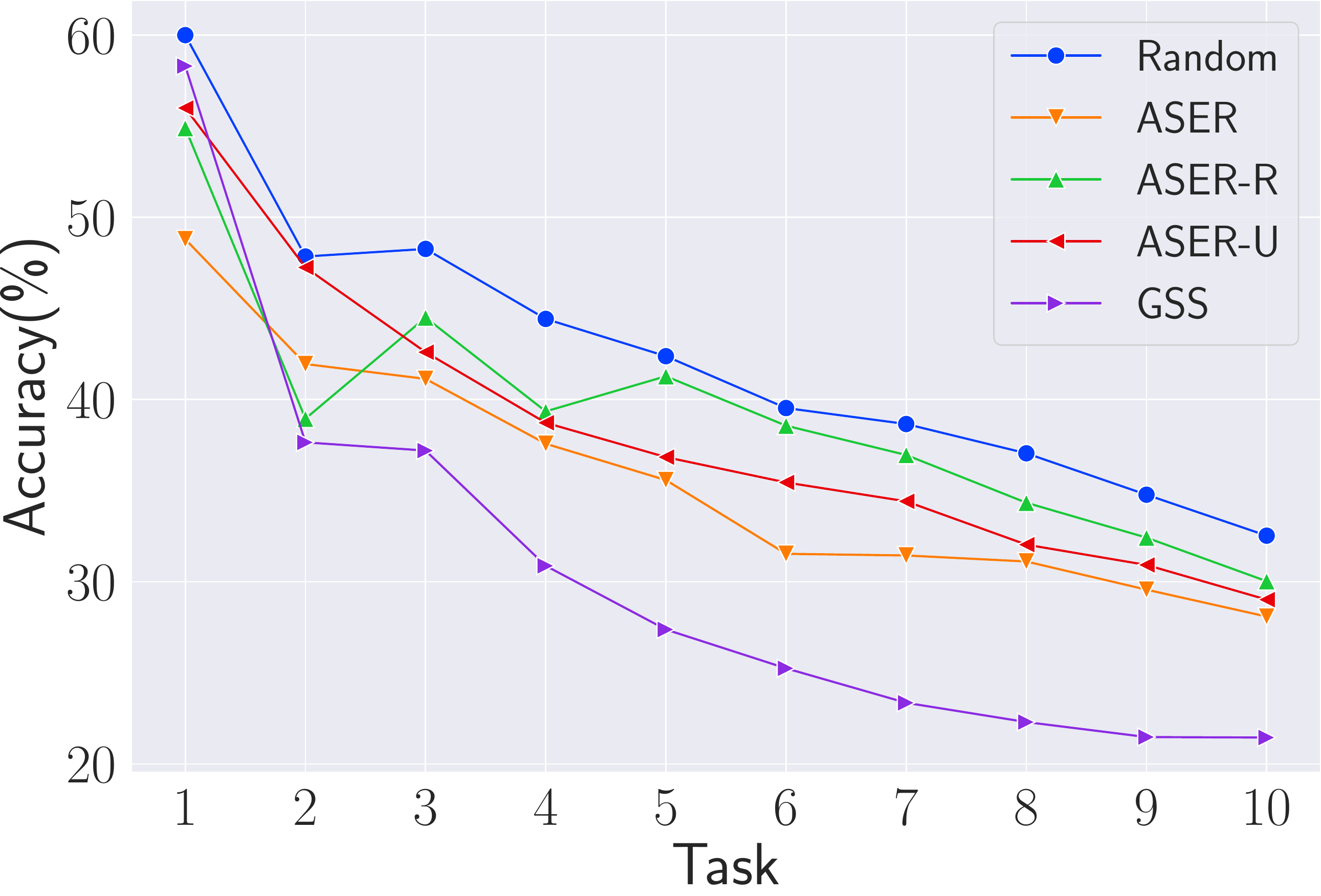}} 
    \subfigure[Impact of temperature]{\includegraphics[width=0.24\textwidth]{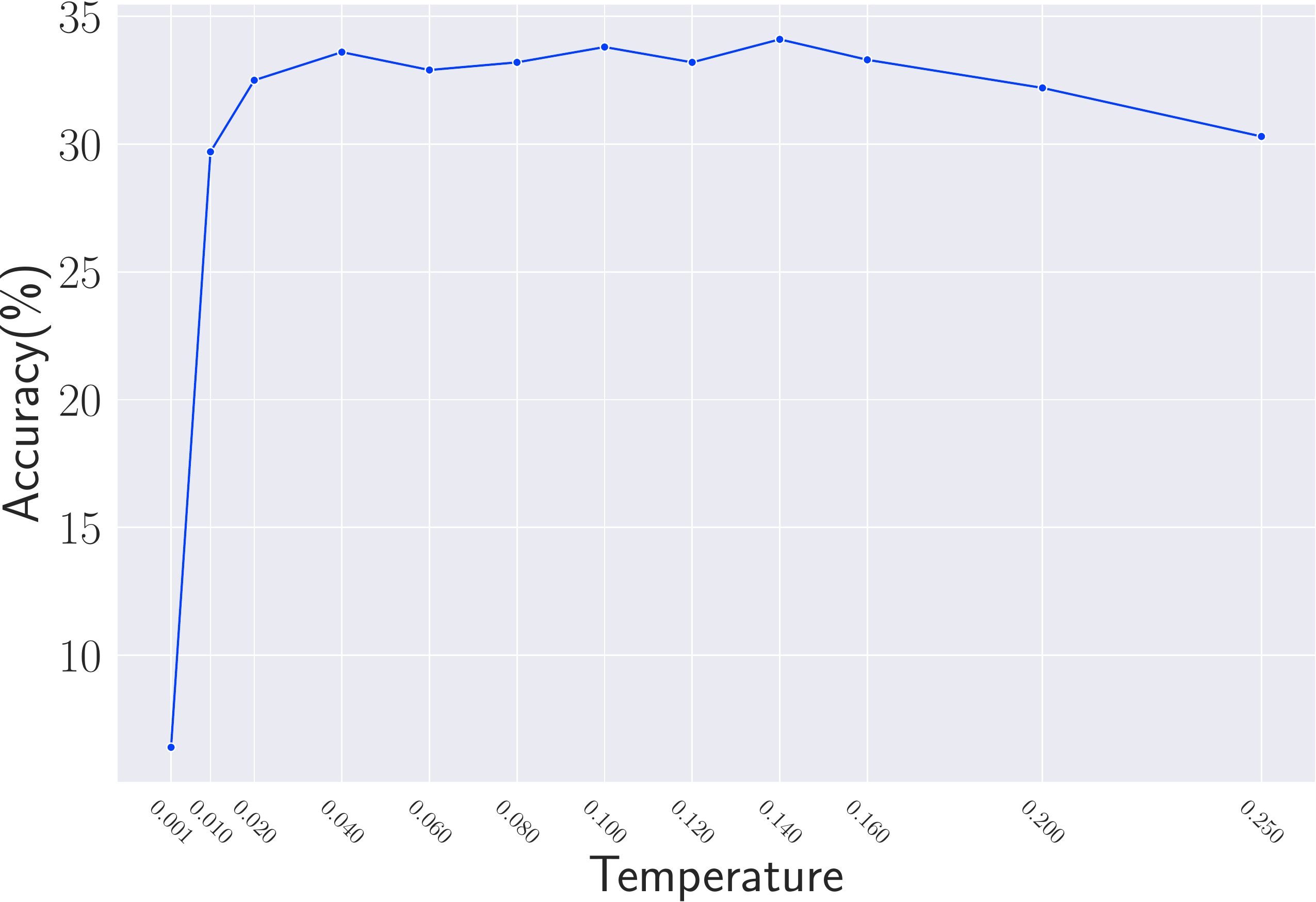}}
    \subfigure[Impact pf projection network]{\includegraphics[width=0.24\textwidth]{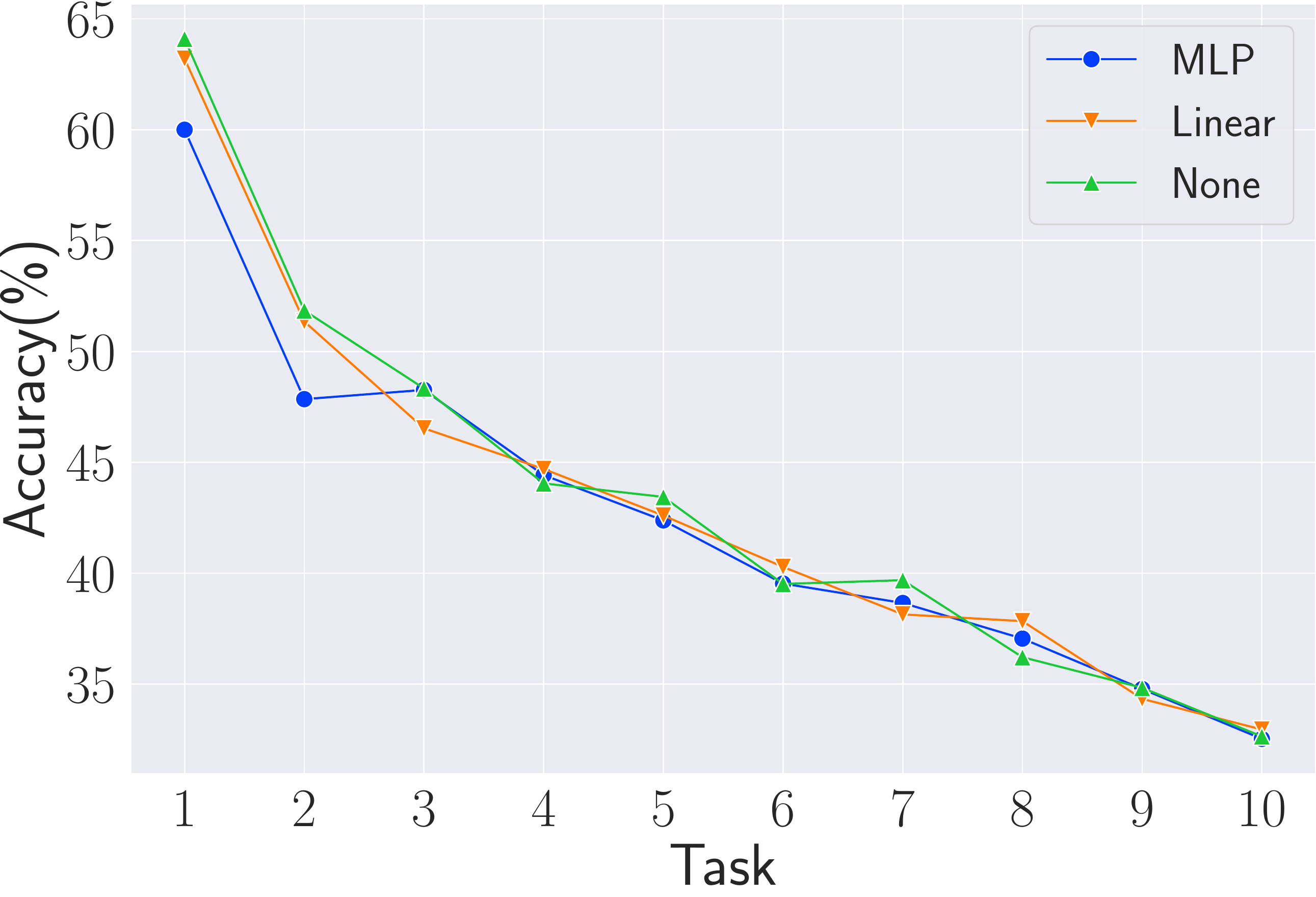}}
    \caption{Average accuracy of SCR with M=2k on CIFAR100 for ablation study. }
    % We study the impact of the memory batch size $B_\mathcal{M}$, the memory buffer retrieval/update methods, the temperature variable $\tau$ and the projection network type.
    \label{fig:ablation}
\end{figure*}

% To test the efficacy of the NCM classifier and our proposed SCR, we compare five methods that employ memory buffers with their variants equipped with the NCM classifier. Moreover, we evaluate the performance of SCR by comparing them with s. We begin by reviewing the benchmark datasets, baselines we compared against and our experiment setting. We then report and analyze the result to validate our approach followed by the ablation study.

\subsection{Evaluation of SCR}
To evaluate the performance of SCR, we compare it with several state-of-the-art CL methods described in Section~\ref{sec:setup}. As we can see in Figure~\ref{fig:line}, SCR consistently outperforms all the compared methods by enormous margins along the whole data streams of three different datasets. Note that all the compared methods on the plots have already been NCM-augmented. Table~\ref{tab:acc_main} shows the detailed comparison of SCR with all the compared methods on different datasets and memory sizes. The last row of the table shows the absolute improvements over the second-best methods.  SCR consistently achieves state-of-the-art results across all settings and outperforms the compared methods by large margins. SCR achieves 35.4\% ($13.3\%\textcolor{red}\uparrow$), 37.8\% ($8.2\%\textcolor{red}\uparrow$) and 65.7\% ($15.4\%\textcolor{red}\uparrow$) respectively in Mini-ImageNet, CIFAR100 and CIFAR10 respectively. The success of SCR comes from (i) the NCM classifier, which has shown impressive performance over the Softmax classifier in Section~\ref{sec: ncm_eval}, and (ii) the contrastive loss, which enables the model to learn more discernible embeddings and provides a solid foundation for the NCM classifier. Moreover, we observe SCR benefits from a large memory buffer in general, as contrastive learning desires more diverse negative samples. For example, SCR achieves 60.2\% relative gain with M=5k (21.1\% (MIR-NCM) $\rightarrow$ 35.4\%), while obtains 35.4\% relative improvement with M=1k (16.6\% (MIR-NCM) $\rightarrow$ 24.1\%). In terms of task-recency bias, we can see in Figure~\ref{fig:confusion} (b) that SCR is clearly much less biased than ER even though a slight bias is still observed. Furthermore, SCR does not sacrifice its computation efficiency, as shown in Figure~\ref{fig:run_time}. Its running time (combined training and inference) is shorter than ASER$_\mu$ and only slightly longer than MIR.

In summary, by evaluating on three standard CL datasets and comparing to the state-of-the-art CL methods, we have strongly demonstrated the effectiveness and efficiency of SCR in overcoming catastrophic forgetting, which brings online CL much closer to its ultimate goal of matching offline training while maintaining a low computation footprint.

\subsection{Ablation Study}
\label{sec: ablation}
In this subsection, we aim to explore the impact of various SCR configurations on its performance.  
 We use SCR with M=2k on CIFAR100 as the study case to analyze the impacts of components of SCR. 
 
 \paragraph{Impact of memory batch size $B_\mathcal{M}$.} Figure~\ref{fig:ablation} (a) shows the impact of the memory batch size. Generally speaking, contrastive learning benefits from larger batch sizes as it means more negative samples~\cite{chen2020simple, khosla2020supervised}. Nevertheless, in online CL, accuracy improvement is more obvious with the increase of $B_\mathcal{M}$ when $B_\mathcal{M}$ is smaller than 200. The performance drops when $B_\mathcal{M}$ continues to increase. We suspect the decrease is due to the overfitting of the memory samples as 500/1,000 are 25\%/50\% of the whole memory buffer in this study case. 

\paragraph{Impact of memory buffer management.} We compare random retrieval + reservoir update(Random), ASER retrieval + update (ASER), ASER retrieval (ASER-R), ASER update (ASER-U) and GSS. As we can see in Figure~\ref{fig:ablation} (b), the random option is much better than GSS and slightly better than others. We observed that some classes have only a few or zero samples in the memory for GSS, which is undesirable for SCR. Although random seems reasonable for the balanced CIFAR100 dataset, when facing imbalanced datasets, combing SCR with other memory management methods may yield better performance~\cite{kim2020imbalanced, chrysakis2020online}.  

\paragraph{Impact of temperature variable $\tau$.}
We can see from Figure~\ref{fig:ablation} (c) that the performance deteriorates when the $\tau$ is too low and too high. SCR with $\tau$ ranging from 0.02 to 0.16 achieves stable results. 

\paragraph{Impact of projection network $Proj(\cdot)$.} We tried Multi-Layer Perceptron (MLP), linear and no projection network (None). Although \cite{chen2020simple} suggests that a nonlinear projection network improves the representation quality, we find that the choice of projection network is insignificant in online CL as shown in Figure~\ref{fig:ablation} (d).

\section{Conclusion}
In this paper, we first demonstrated that the NCM classifier is a simple yet effective substitute for the Softmax classifier in the online CL. It resolves several deficiencies of the Softmax classifier and shows considerable and consistent performance gains across a variety of CL methods. Based on these results, we advocate using the NCM classifier instead of the commonly used Softmax classifier for future study of CL methods. Moreover, to leverage the NCM classifier more effectively, we proposed SCR that explicitly encourages samples from the same class to cluster tightly in embedding space while pushing samples of different classes further apart during experience replay-based training. 

Empirically, we observe that our proposed SCR substantially reduces catastrophic forgetting in comparison to state-of-the-art CL methods and outperforms them all by a significant margin on various datasets and memory settings.  In summary, leveraging a simple randomized experience replay method while using a supervised contrastive loss (in place of cross-entropy) combined with an NCM classifier bring us closer to realizing the ultimate goal of continual learning to perform as well as offline training methods.

{
\newpage
\bibliographystyle{ieee_fullname}
\bibliography{egbib}
}

\end{document}